\documentclass[]{spie}  

\usepackage{amsmath,amsfonts,amssymb}
\usepackage{graphicx}
\usepackage[colorlinks=true, allcolors=blue]{hyperref}

\title{Depth Reconstruction and Computer-Aided Polyp Detection in Optical Colonoscopy Video Frames**}

\author[]{Saad Nadeem}
\author[]{Arie Kaufman}
\affil[]{Department of Computer Science, Stony Brook University, Stony Brook, NY, 11794, USA}

\authorinfo{Preprint for accepted SPIE Medical Imaging Conference 2016 manuscript. DOI: 10.1117/12.2216996
\newline
\textbf{**The title has been modified to highlight the contributions more clearly.}
Please cite as follows:\\
Saad Nadeem and Arie Kaufman. ``Computer-Aided Detection of Polyps in Optical Colonoscopy Images.'' SPIE Medical Imaging, 2016.}

\begin{document}
\maketitle

\begin{abstract}
We present a computer-aided detection algorithm for polyps in optical colonoscopy images. Polyps are the precursors to colon cancer. In the US alone, more than 14 million optical colonoscopies are performed every year, mostly to screen for polyps. Optical colonoscopy has been shown to have an approximately 25\% polyp miss rate due to the convoluted folds and bends present in the colon. In this work, we present an automatic detection algorithm to detect these polyps in the optical colonoscopy images. We use a machine learning algorithm to infer a depth map for a given optical colonoscopy image and then use a detailed pre-built polyp profile to detect and delineate the boundaries of polyps in this given image. We have achieved the best recall of 84.0\% and the best specificity value of 83.4\%.
\end{abstract}


\section{INTRODUCTION}
\label{sec:intro}  

Colorectal cancer (CRC) is the third most frequently diagnosed cancer worldwide, and the fourth leading cause of cancer deaths with 700,000 deaths worldwide per year \cite{cancerincidence,cancermortality}. It is expected to become more common as more people adopt Western diets and lifestyles, which are implicated as risk factors. Early detection and removal of polyps, the precursors to colorectal cancer, is critical in extending the life of a patient with potential colon cancer. Optical colonoscopy (OC) is the most prevalent tool for the detection and removal of these polyps; more than 14 million optical colonoscopies are performed every year in the US \cite{seeff:2004}. However, due to the convoluted structure of the colon, the polyp miss rate in OC stands at approximately 25\%. For this purpose, automatic detection and delineation of polyps in post-procedure colonoscopy video frames can be helpful in confirming the diagnosis and in documenting the procedure. A thorough documentation of the polyps, if found, can be especially helpful for prognosis (comparison with follow-up procedures) purposes.

Virtual colonoscopy (VC) is a non-invasive alternative to optical colonoscopy in which computed tomography (CT) scans of a patient's abdominal region are taken and a three dimensional image is reconstructed from these scans for colon examination of polyps. In other words, VC 3D reconstructed colon can be considered a static snapshot of a real colon, observed during OC with additional three dimensional geometric information, such as depth and normals. This three dimensional information allows features like maximum, minimum, and mean of principal curvatures of surface \cite{summers:2001}, 3D geometric features \cite{gokturk:2001}, shape in the vicinity of a voxel \cite{Yoshida:2002}, and lines of curvatures \cite{Zhao:2006} to be used for effective detection of polyps in VC. Polyps vary in shape, size and complex surroundings and thus additional 3D geometric features, like in VC, can be of critical importance in precise and accurate detection of polyps.

The lack of additional 3D geometric information (for effective detection of polyps) in 2D OC images has spurred efforts for extracting this 3D information using various computer vision techniques. However, OC presents a highly challenging environment for these general computer vision techniques to work effectively. Since, in OC, the endoscope has a monocular lens camera and the movement of the camera is highly unpredictable, the promising shape-from-stereo \cite{cohen:1989,scharstein:2002} approaches have been rendered useless. Similarly the inconsistent texture patterns on the colon wall are not conducive to shape-from-texture \cite{forsyth:2002} approaches either. Shape-from-motion (SfM) techniques \cite{Koppel:2007} have shown some promise under the invalid assumption that the colon is a rigid object and thus the shape does not change between images. Shape-from-shading (SfS) methods \cite{Deguchi:1996}, on the other hand, only work for OC images that show colon wall or surface (mucosa) and fail when exposed to OC images with endoluminal view. This is because SfS expresses the lumen as a far surface rather than a hole. Kaufman et al. \cite{Kaufman:2008} combined SfS and SfM to reconstruct partial surfaces but only for colon surface view because of the inherent limitations of the two techniques used.

The limitations of these general computer vision techniques in the context of OC have led to feature-specific approaches. Zhou et al. \cite{zhou:2008} use optical flow based method to reconstruct small colon segments under the invalid assumptions that the neighboring folds in an image are not occluded and that the colon fold contours are circular in nature; partial occlusion of folds is typical and the transverse colon segments have characteristic triangular fold contours. Hong et al. \cite{Hong:2014}, on the other hand, used estimated colon folds and the brightness intensity of selected pixels around the fold contour, while assuming certain values for intrinsic camera parameters, to estimate the depth from the camera. This technique can be used to create a colon segment but with the limitation that only the endoluminal view with at least one colon fold clearly outlined is needed.

In contrast to these approaches, we estimate the 3D geometric feature (depth in our case) from an OC image without placing any assumptions or conditions on the viewing angle, video frame characteristics or the intrinsic parameters of the camera and use this 3D depth feature to automatically detect and delineate the boundaries of polyps, if found, in a given OC image.

\section{COMPUTER-AIDED DETECTION OF POLYPS}
Our computer-aided detection pipeline is as follows: (1) Create a general RGB-Depth (RGB images and their corresponding Depth maps) dictionary from VC datasets; (2) Compute the depth map representation of a given input OC image using the RGB-Depth dictionary; (3) Detect and delineate the boundaries of polyps, if found, in the depth map representation of the OC image using our database of VC polyps in the depth map representation.

The novelty of our work extends from the fact that we use VC data to predict depth in OC images. It is hard to use machine learning algorithms with OC data since it is not feasible to build the dictionary with OC images and precise depth information. However, since the VC data is a static snapshot of the corresponding OC data, it can be effectively used to approximate certain geometric features, such as depth. In this work, we use the depth data retrieved from the 3D VC reconstruction to approximate depth in OC images.

\subsection{RGB-Depth Dictionary}
\label{sec:dictionary}

\setlength{\tabcolsep}{2.2pt}
\begin{figure}[ht!]
\begin{center}
\begin{tabular}{cc|cc|cc|cc}
RGB Image & Depth Map & RGB Image & Depth Map & RGB Image & Depth Map & RGB Image & Depth Map\\
\includegraphics[height=0.065\textwidth]{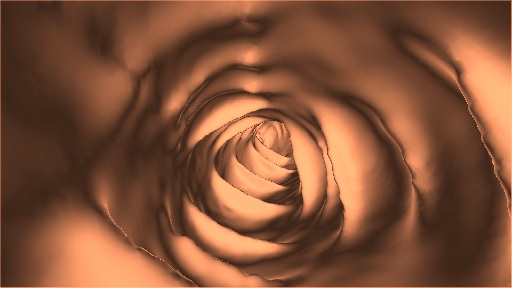}&
\includegraphics[height=0.065\textwidth]{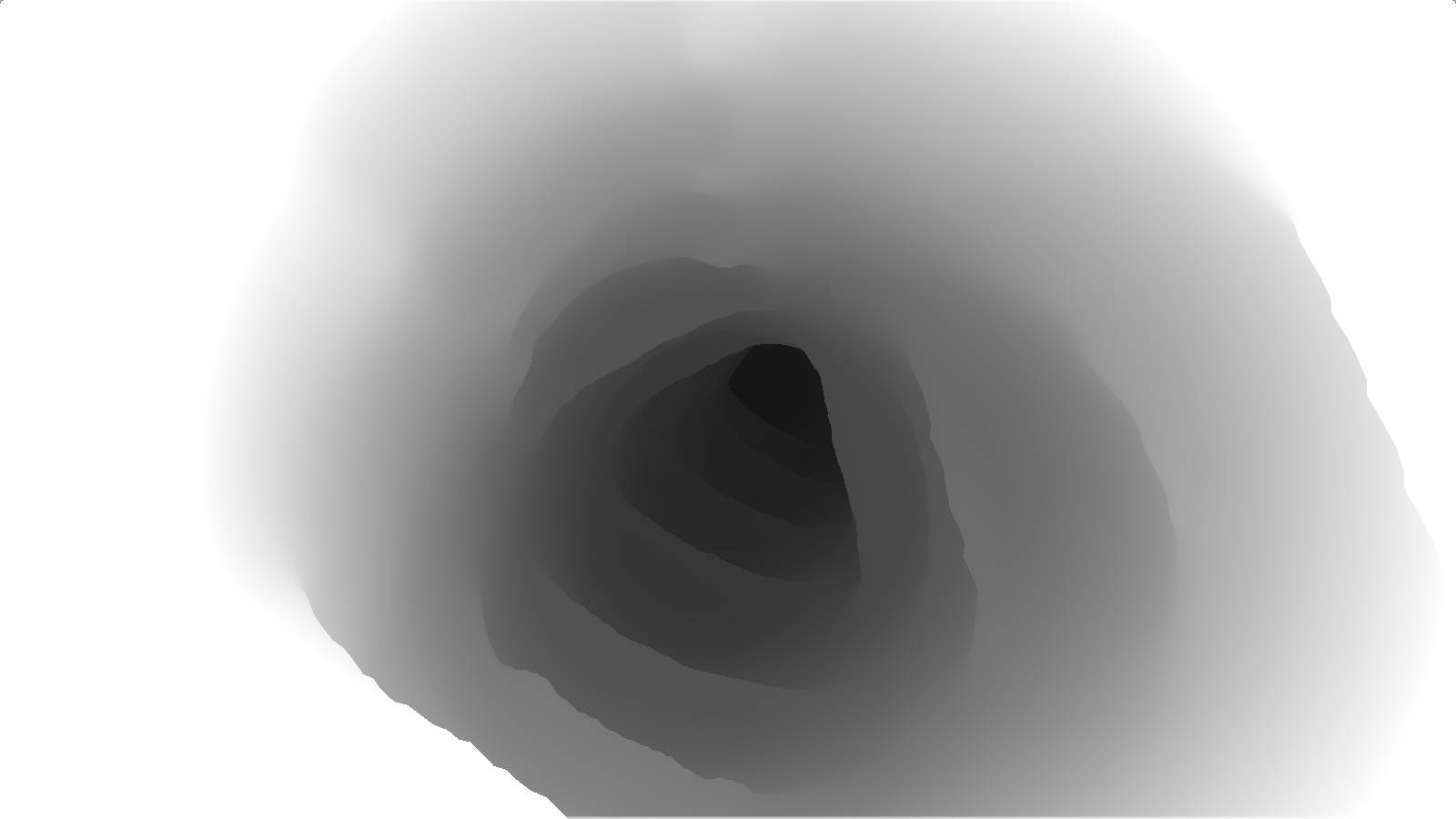}&
\includegraphics[height=0.065\textwidth]{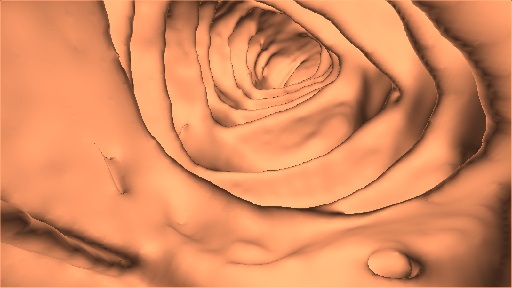}&
\includegraphics[height=0.065\textwidth]{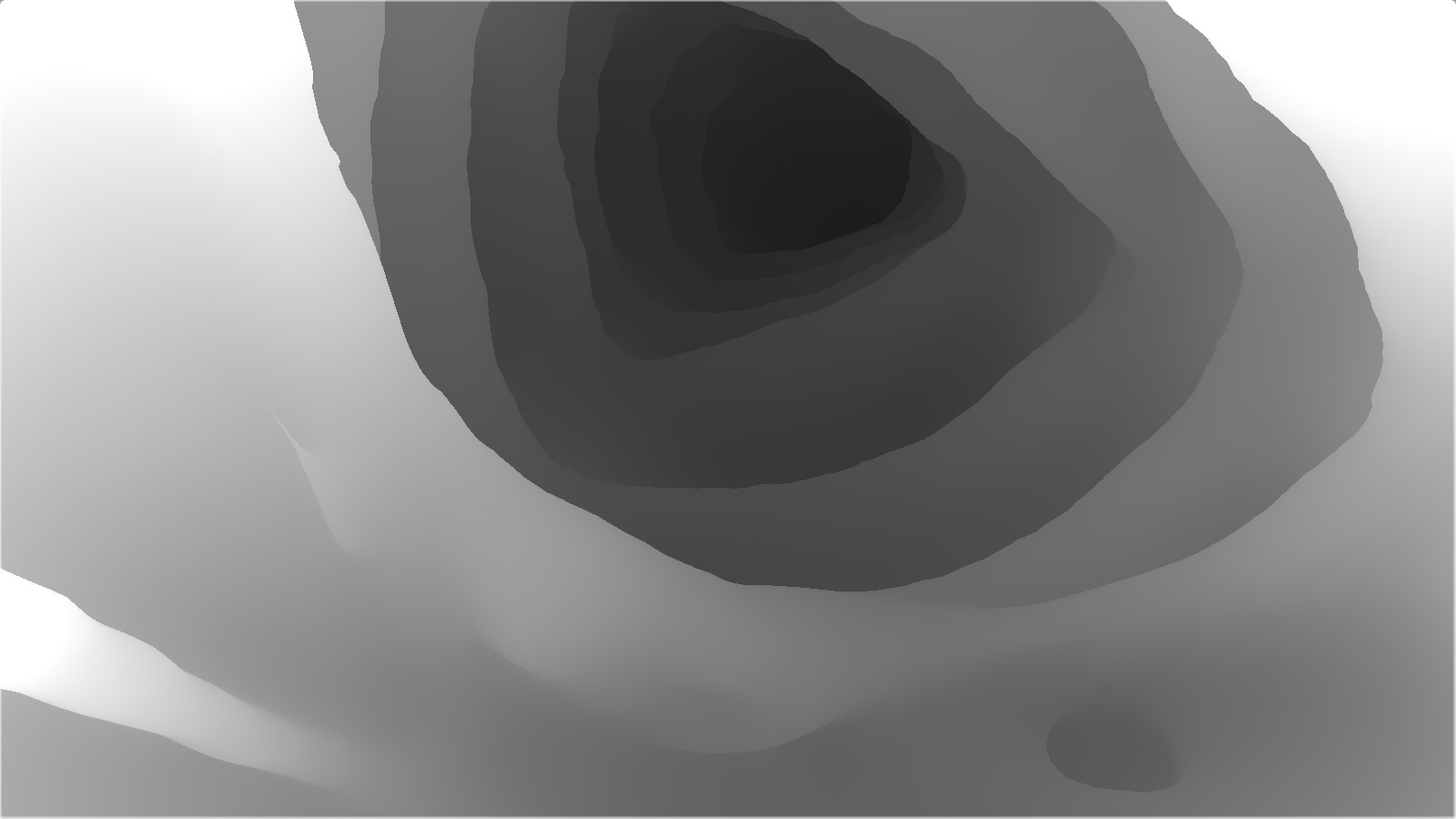}&
\includegraphics[height=0.065\textwidth]{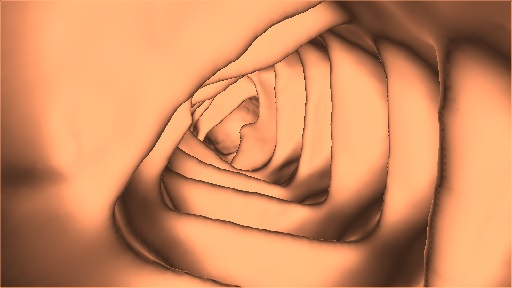}&
\includegraphics[height=0.065\textwidth]{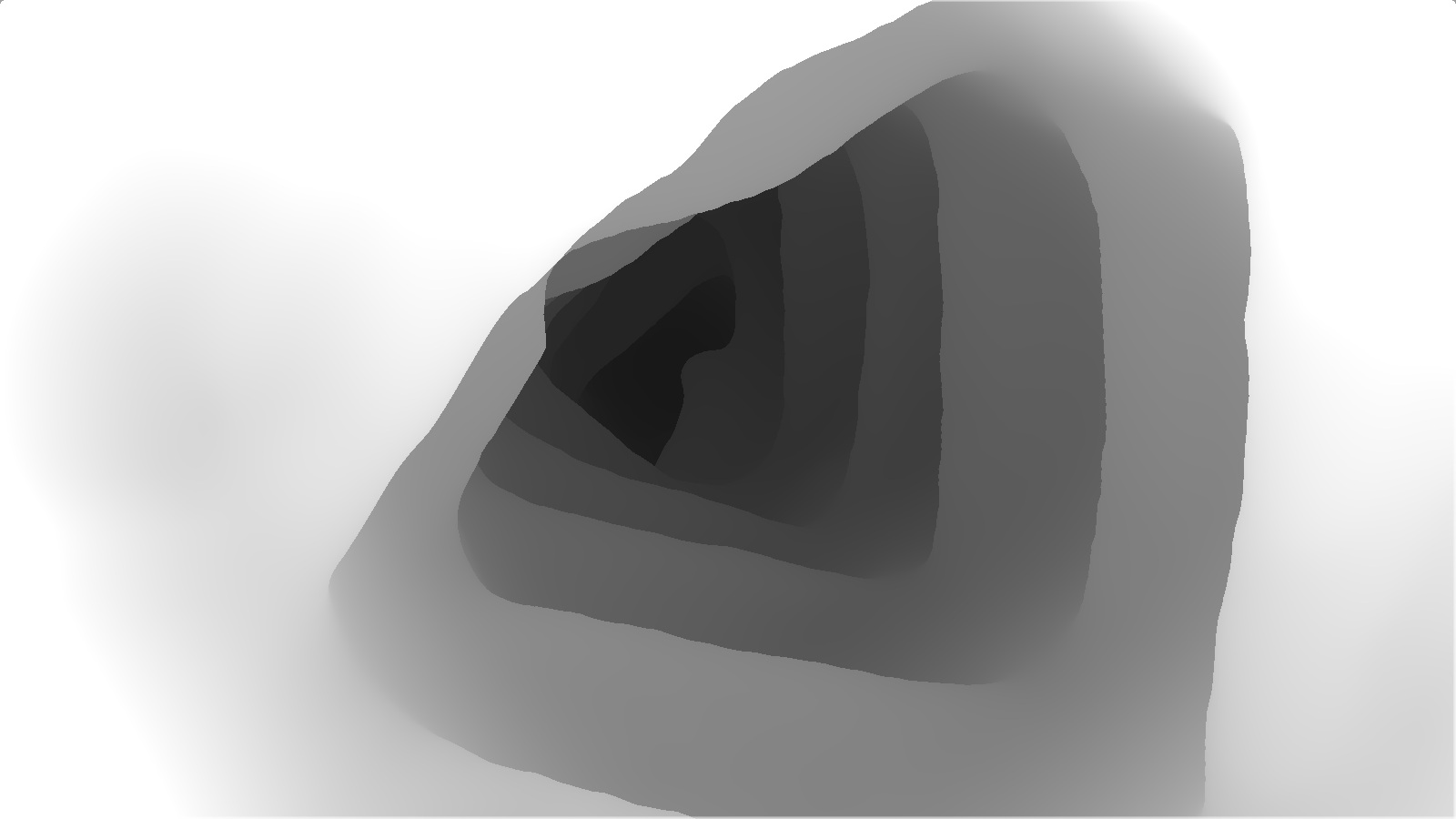}&
\includegraphics[height=0.065\textwidth]{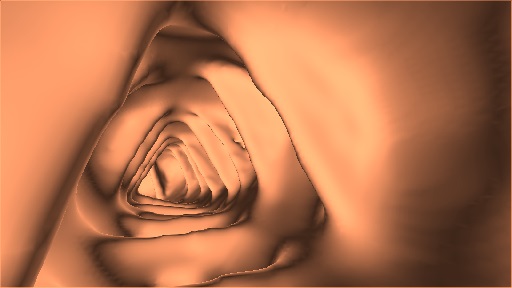}&
\includegraphics[height=0.065\textwidth]{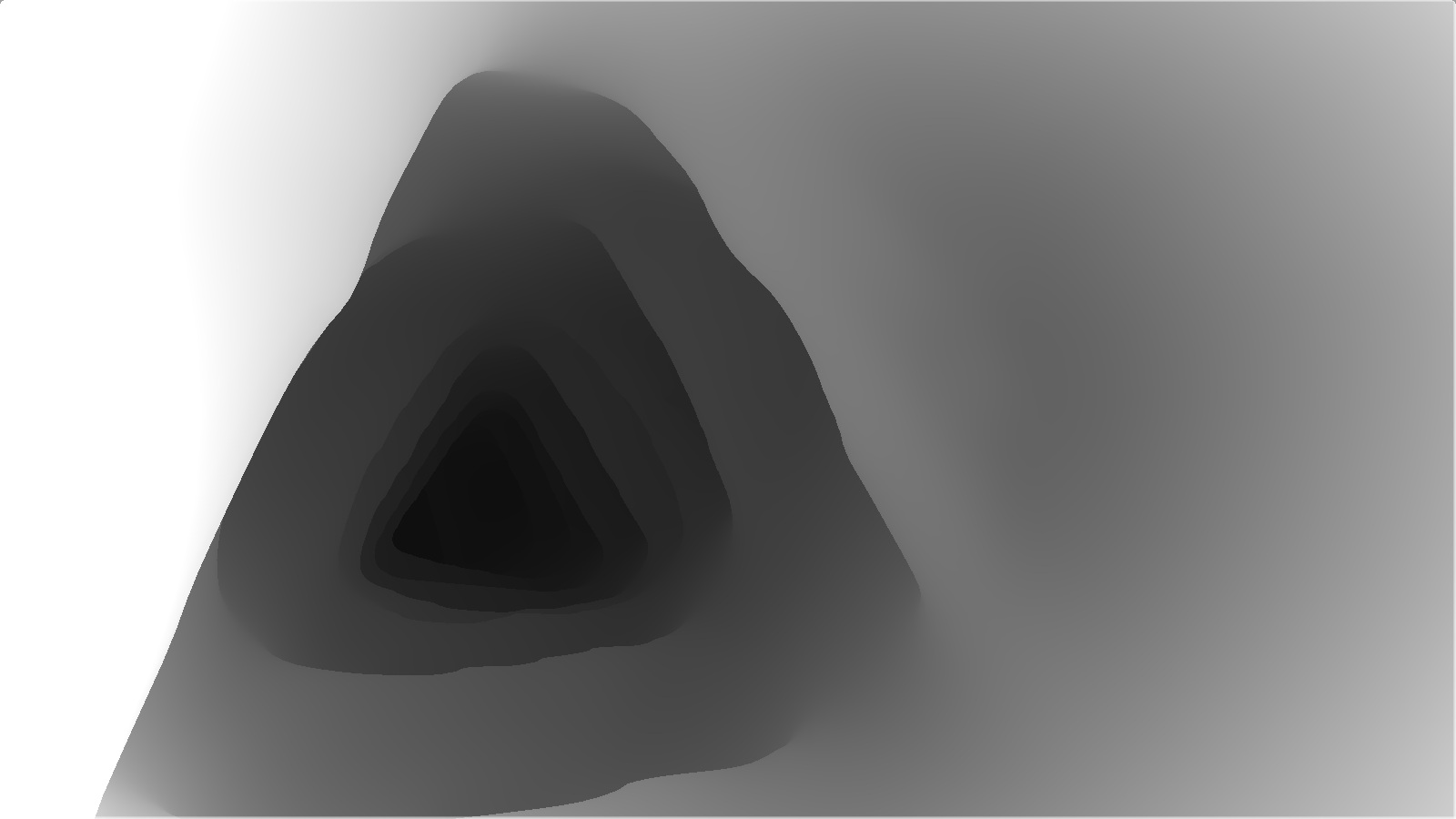}\\
\includegraphics[height=0.065\textwidth]{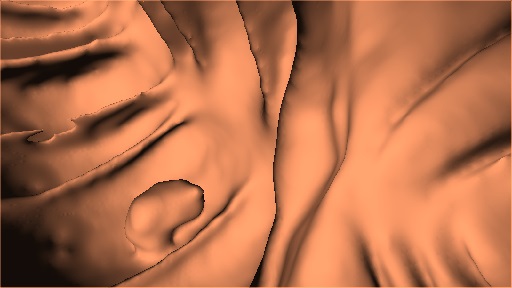}&
\includegraphics[height=0.065\textwidth]{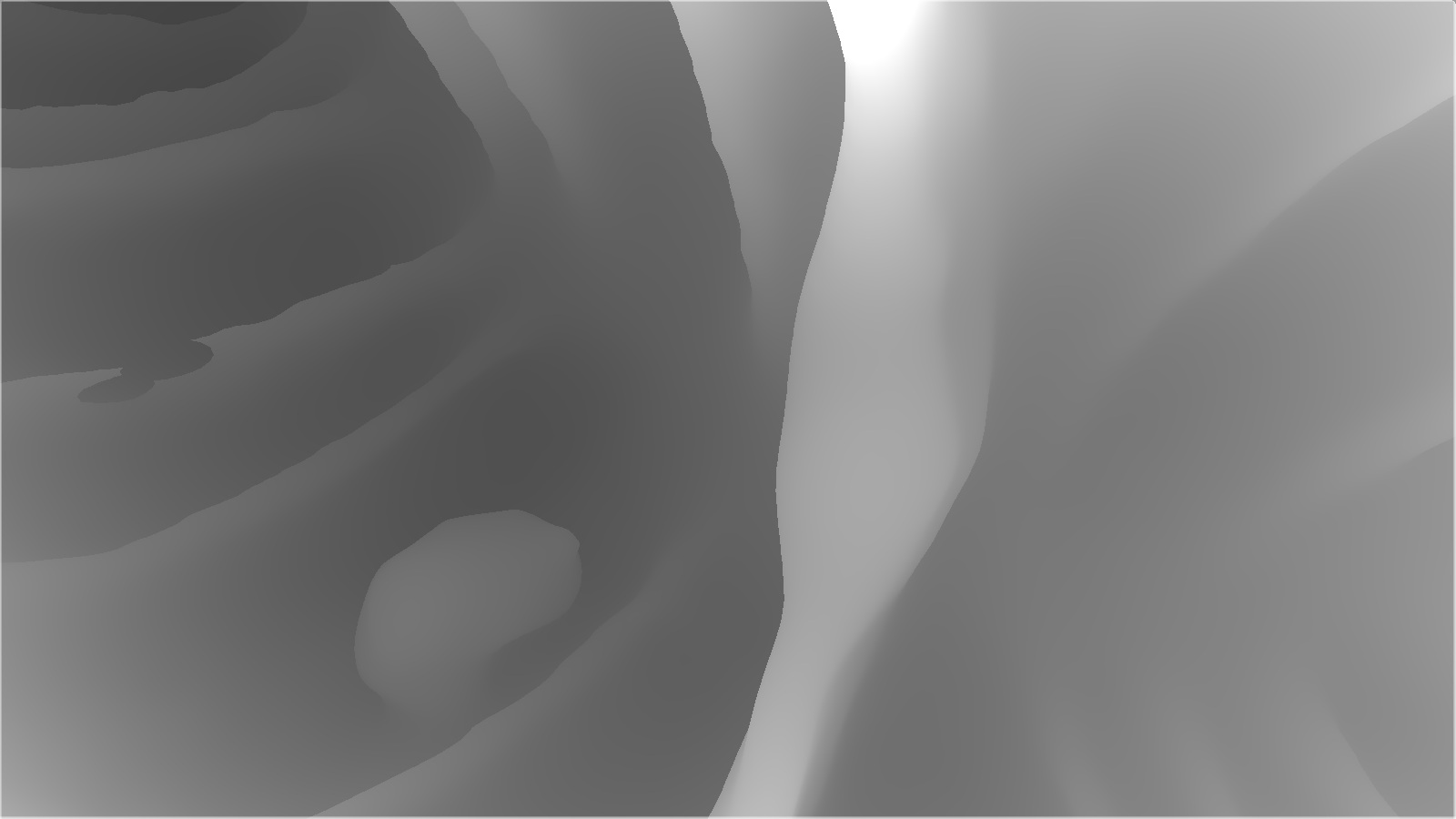}&
\includegraphics[height=0.065\textwidth]{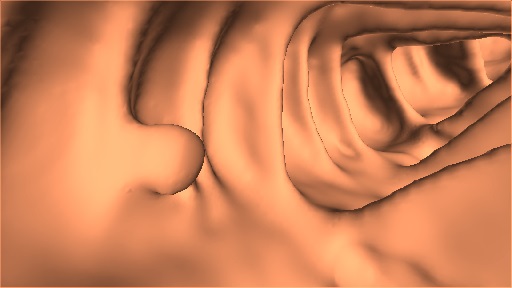}&
\includegraphics[height=0.065\textwidth]{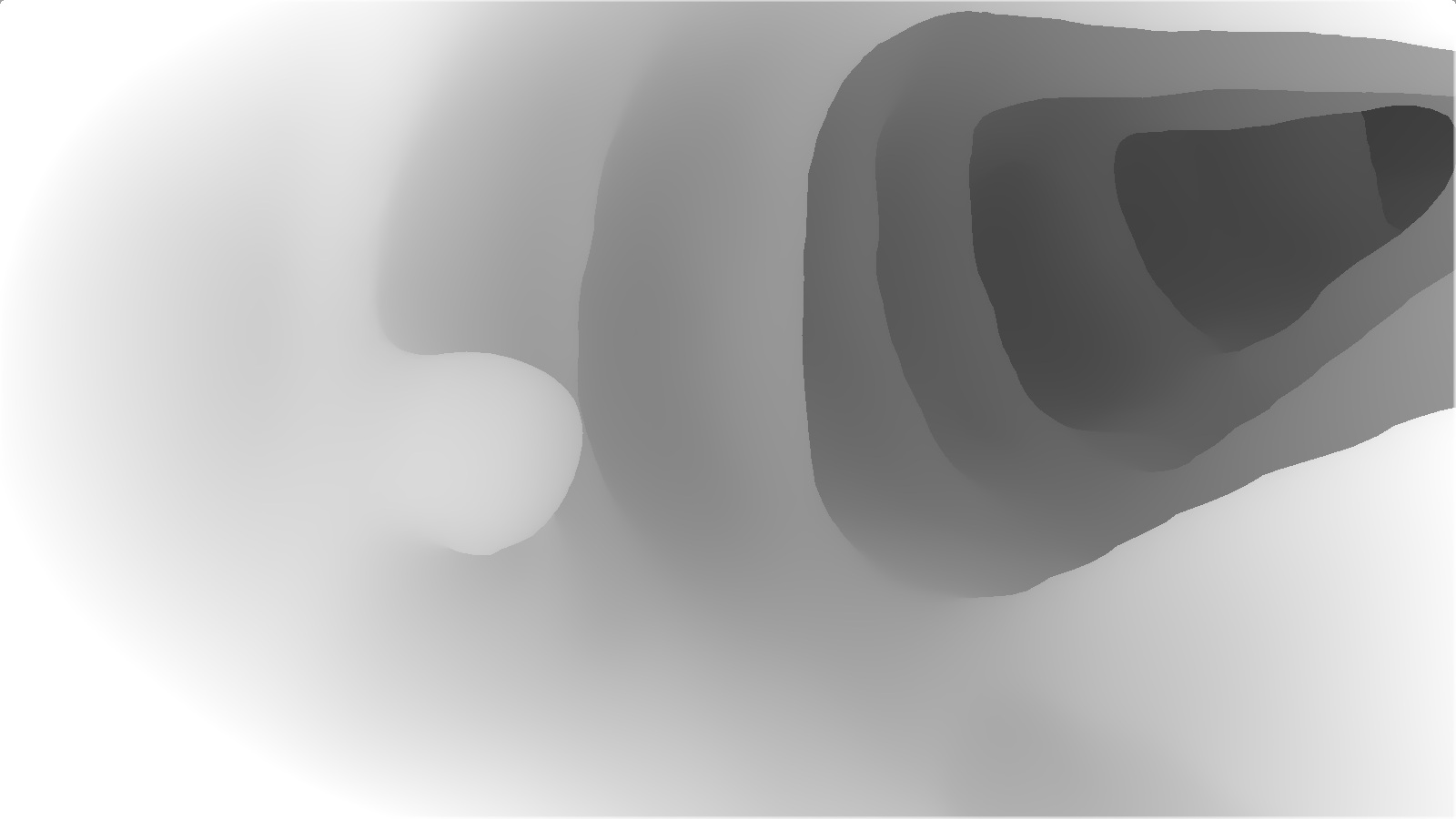}&
\includegraphics[height=0.065\textwidth]{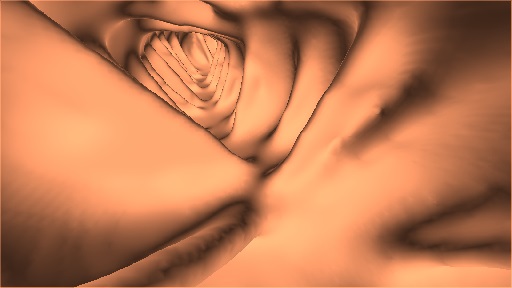}&
\includegraphics[height=0.065\textwidth]{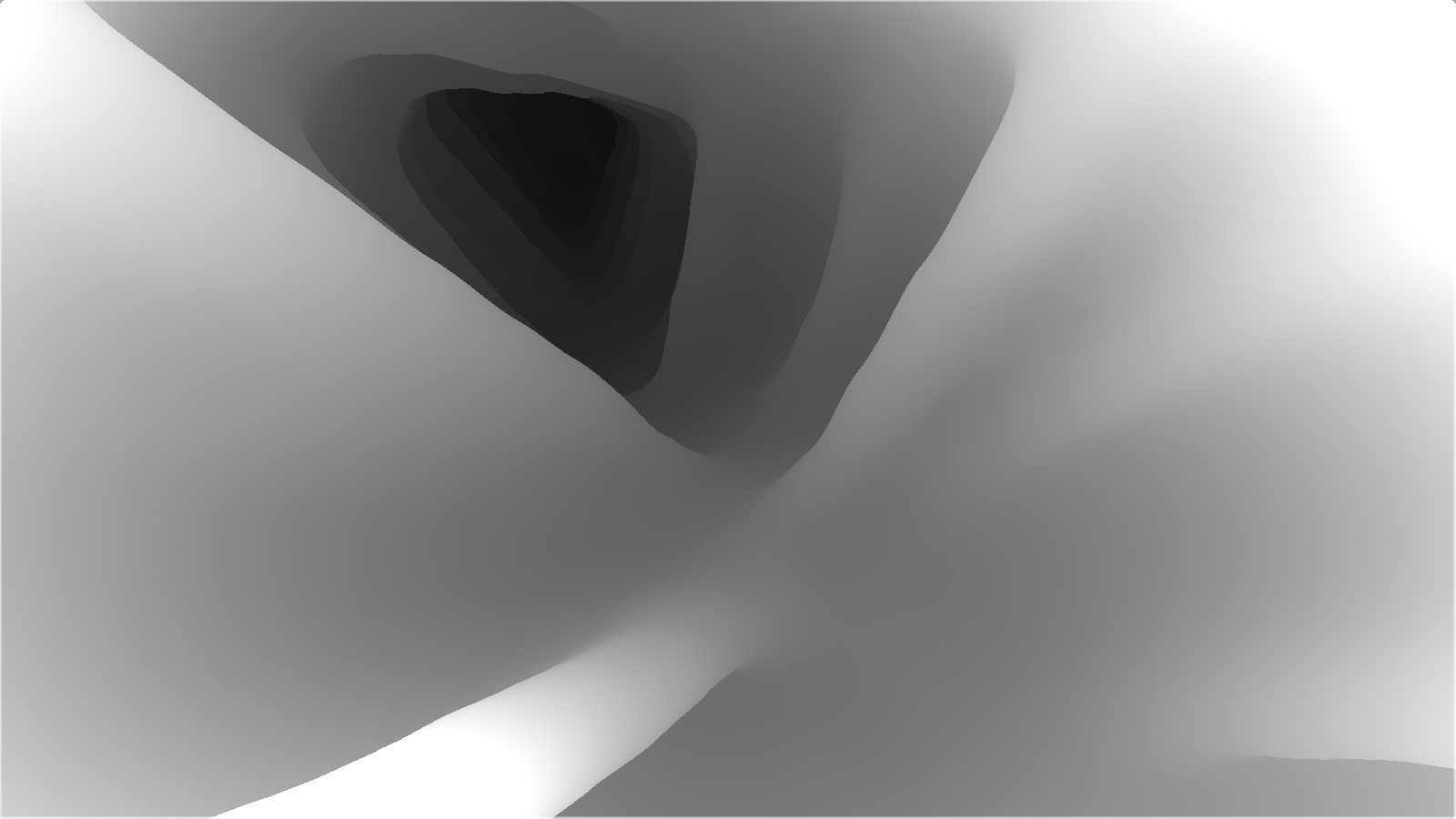}&
\includegraphics[height=0.065\textwidth]{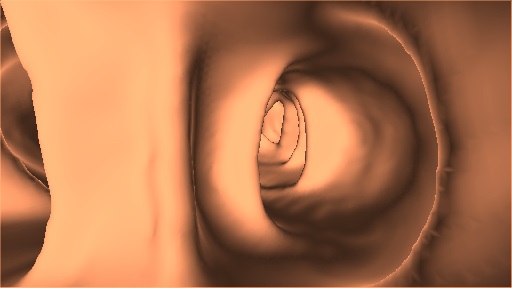}&
\includegraphics[height=0.065\textwidth]{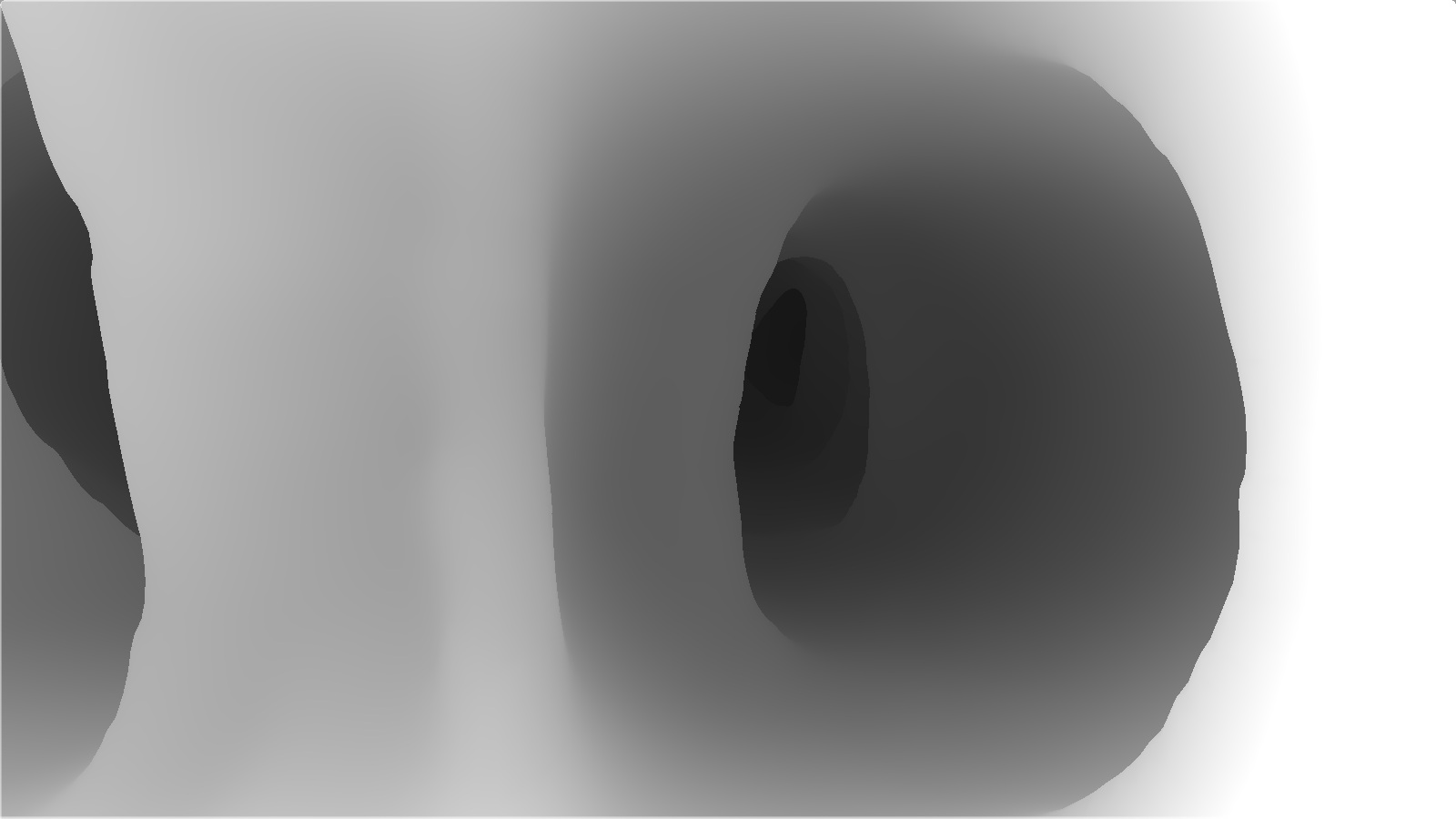}\\
\includegraphics[height=0.065\textwidth]{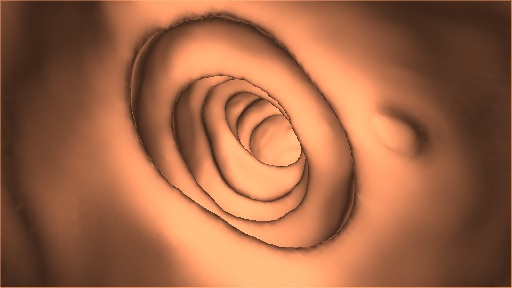}&
\includegraphics[height=0.065\textwidth]{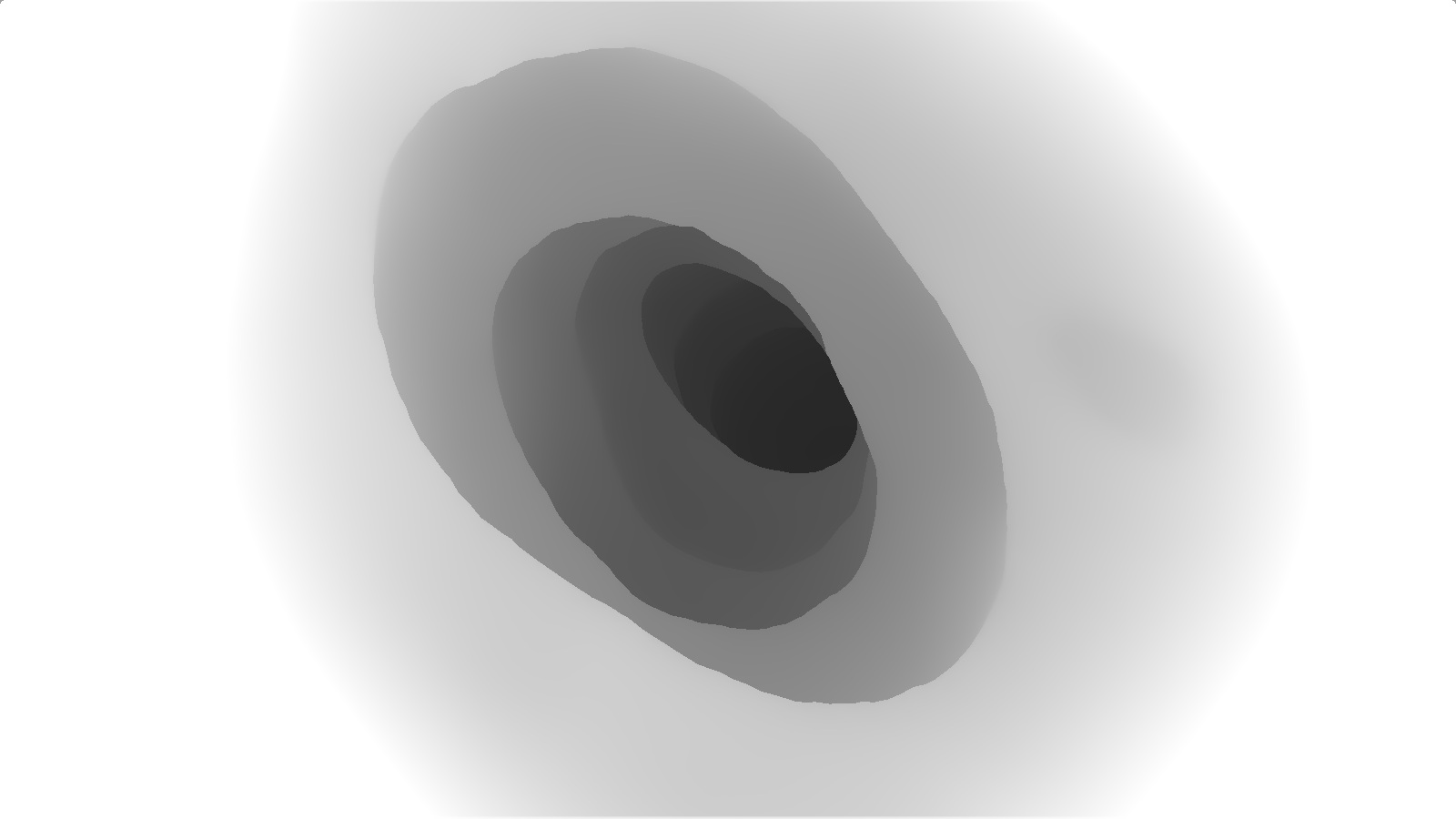}&
\includegraphics[height=0.065\textwidth]{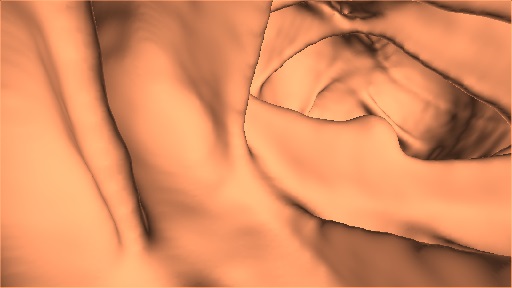}&
\includegraphics[height=0.065\textwidth]{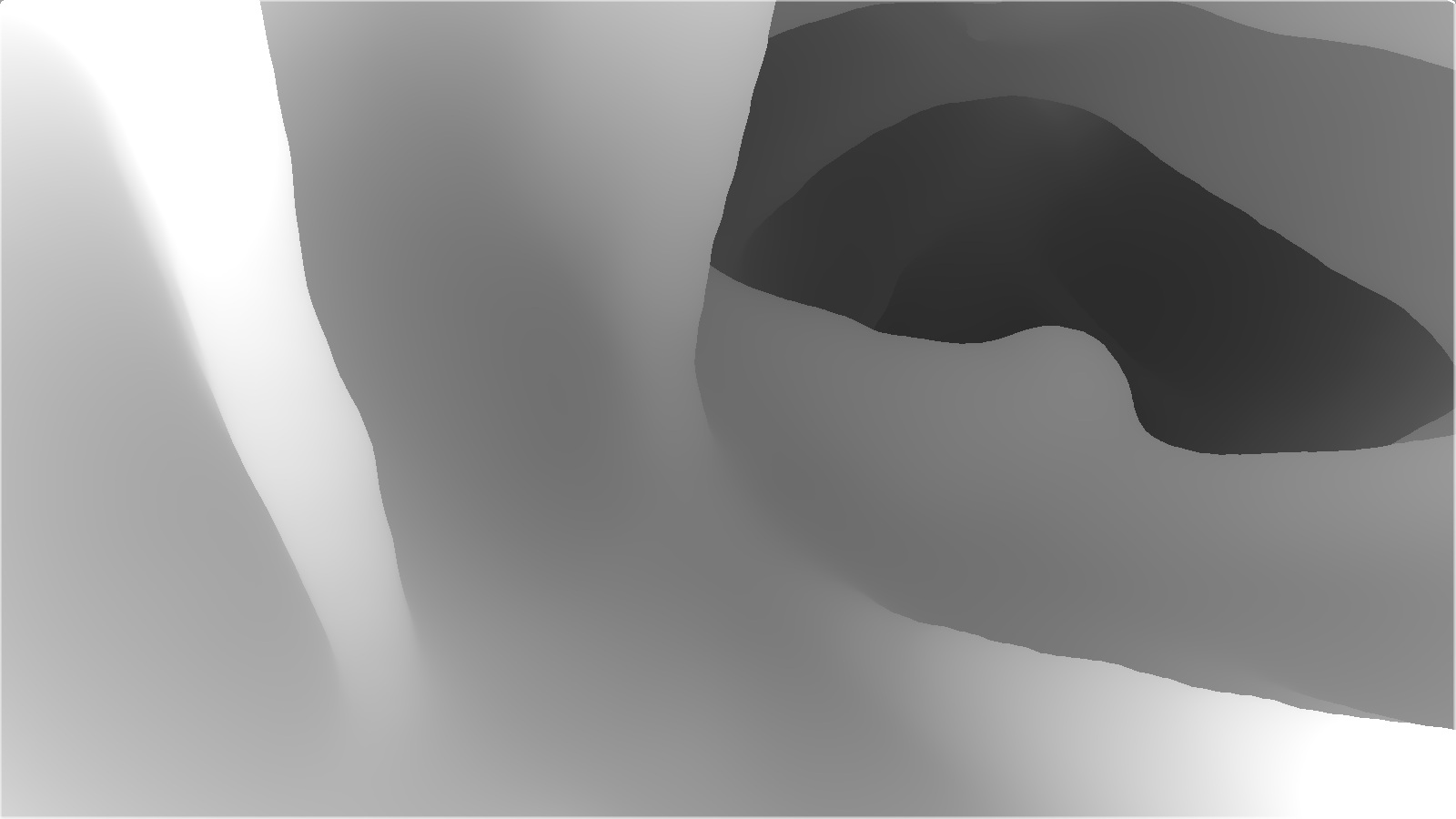}&
\includegraphics[height=0.065\textwidth]{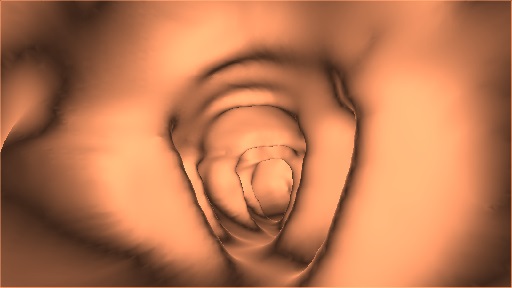}&
\includegraphics[height=0.065\textwidth]{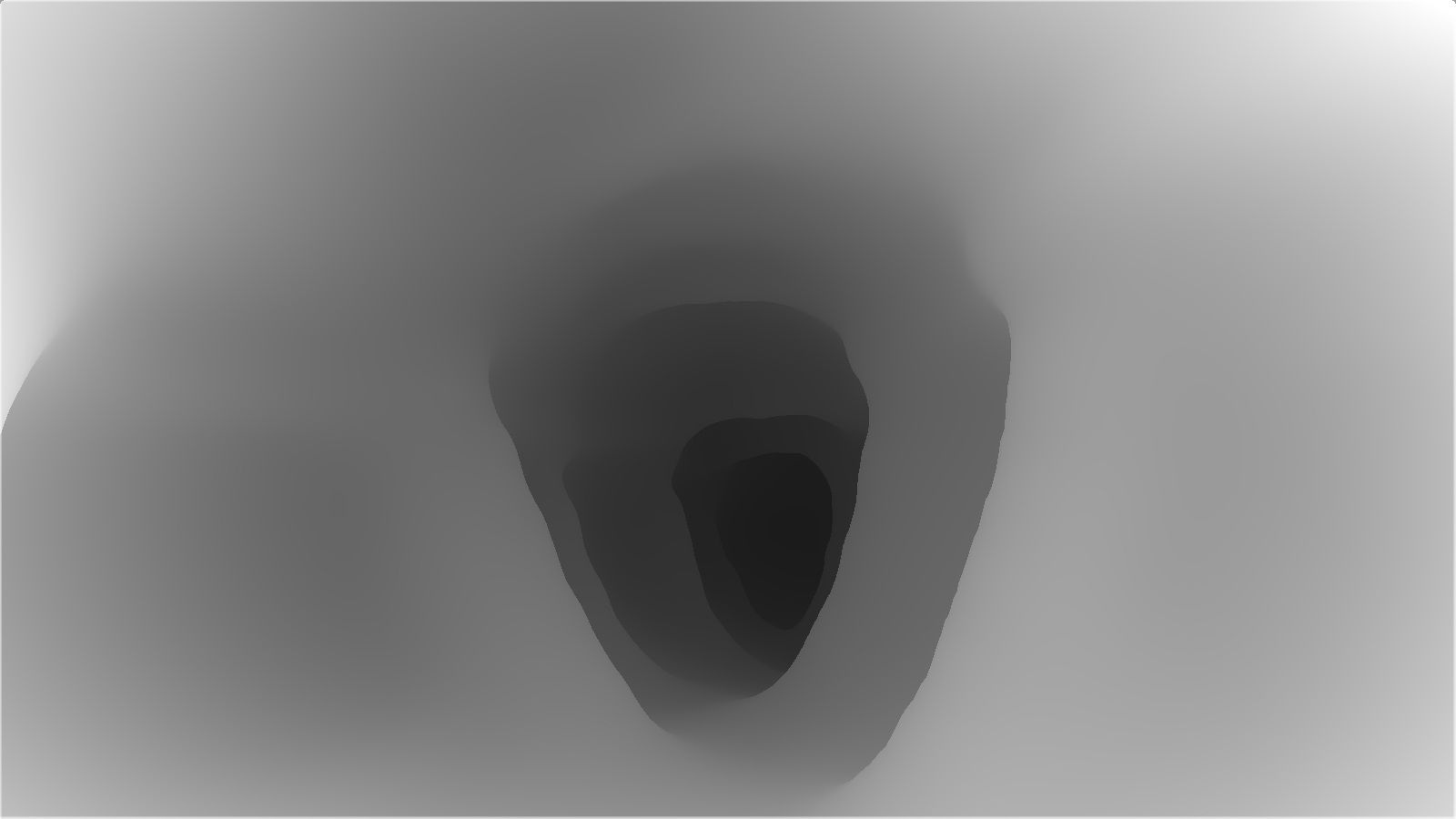}&
\includegraphics[height=0.065\textwidth]{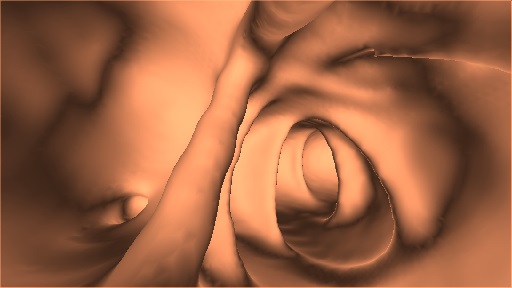}&
\includegraphics[height=0.065\textwidth]{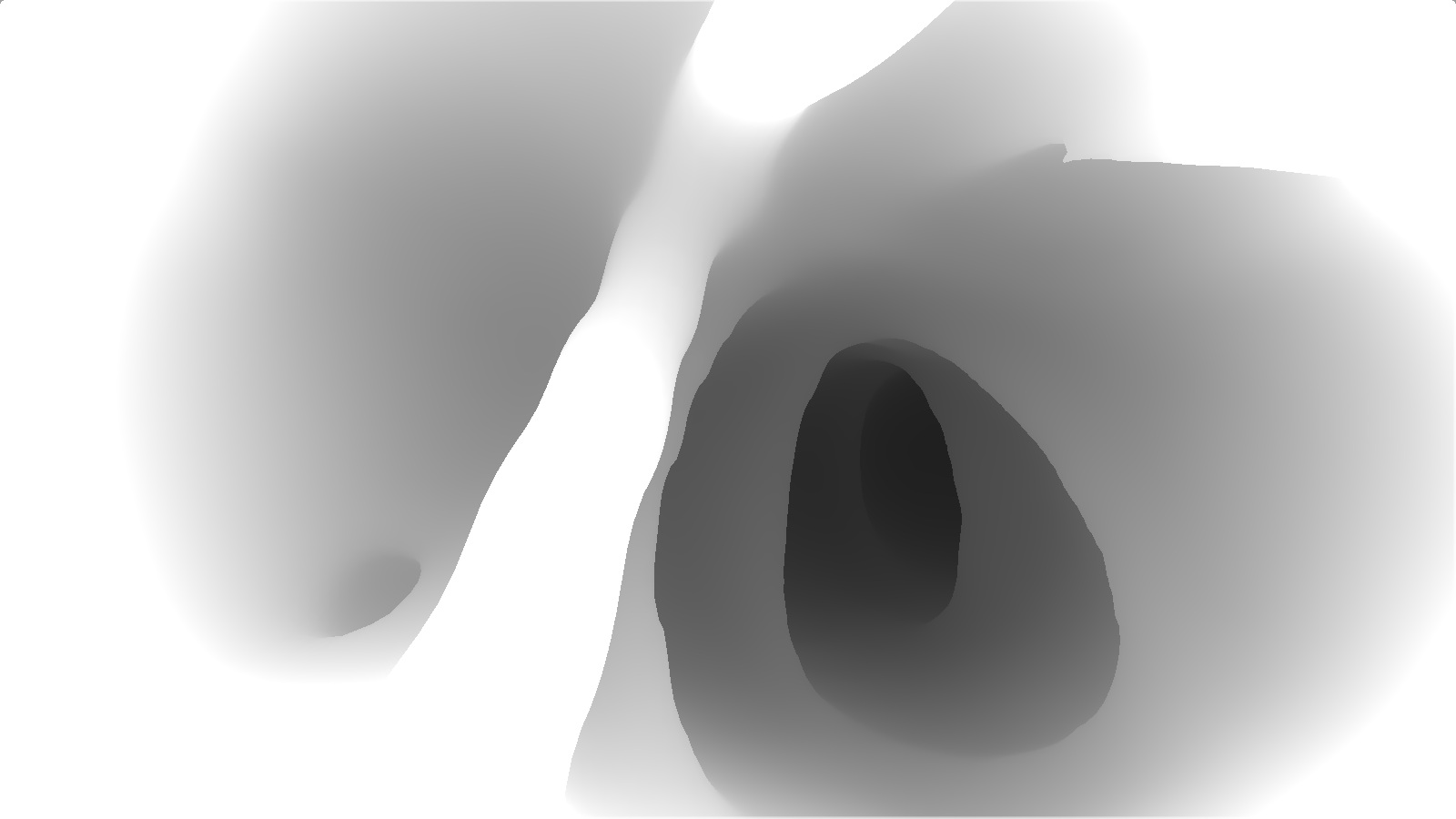}\\
\includegraphics[height=0.065\textwidth]{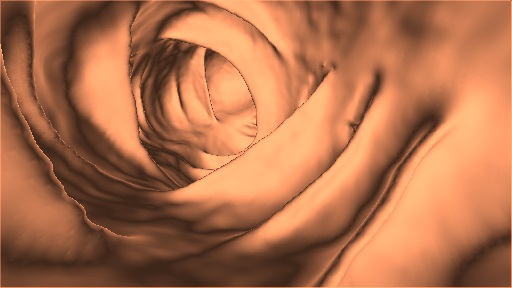}&
\includegraphics[height=0.065\textwidth]{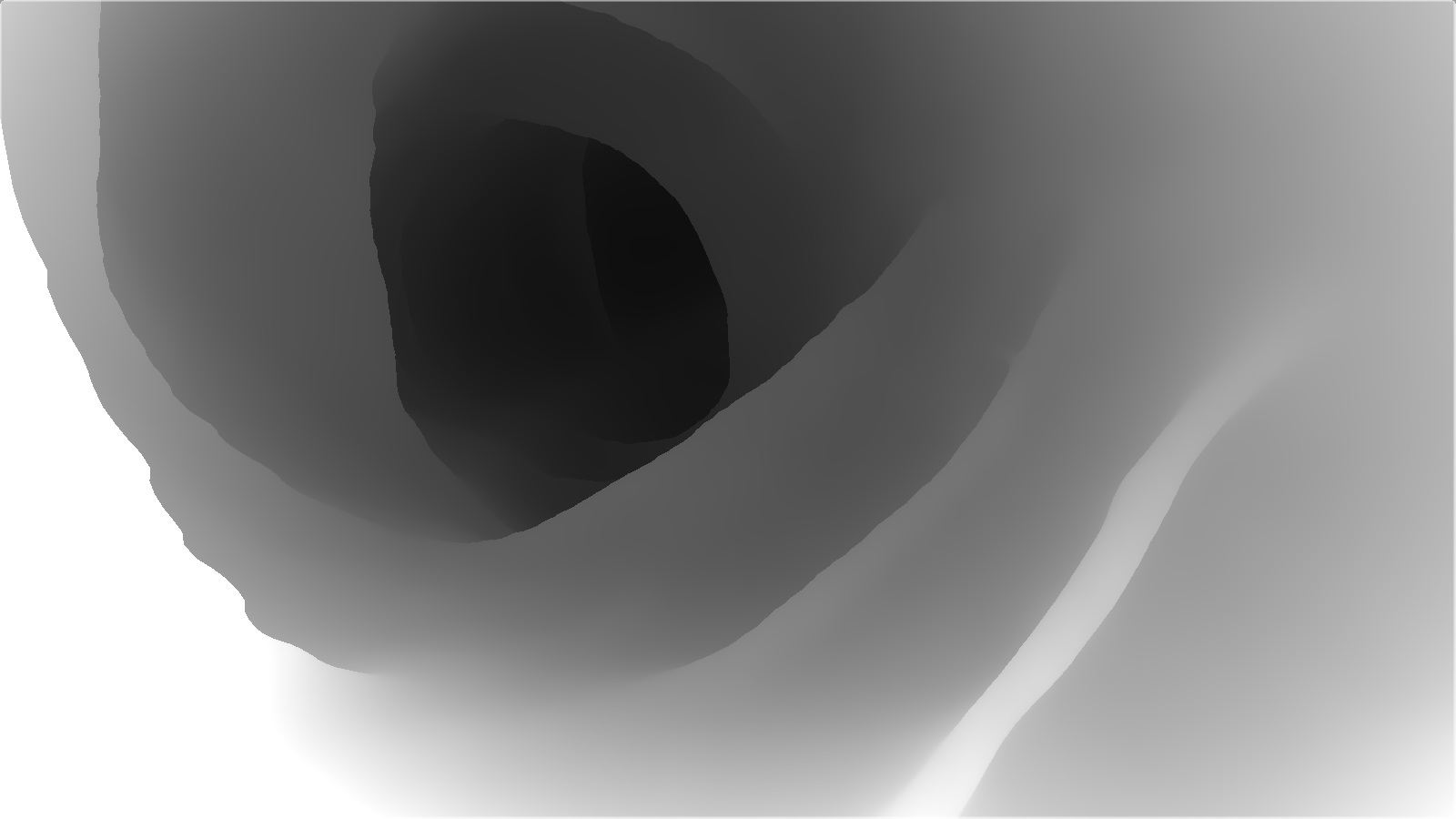}&
\includegraphics[height=0.065\textwidth]{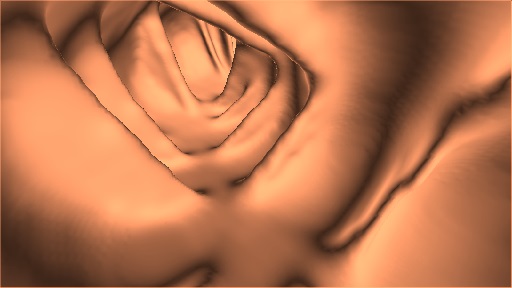}&
\includegraphics[height=0.065\textwidth]{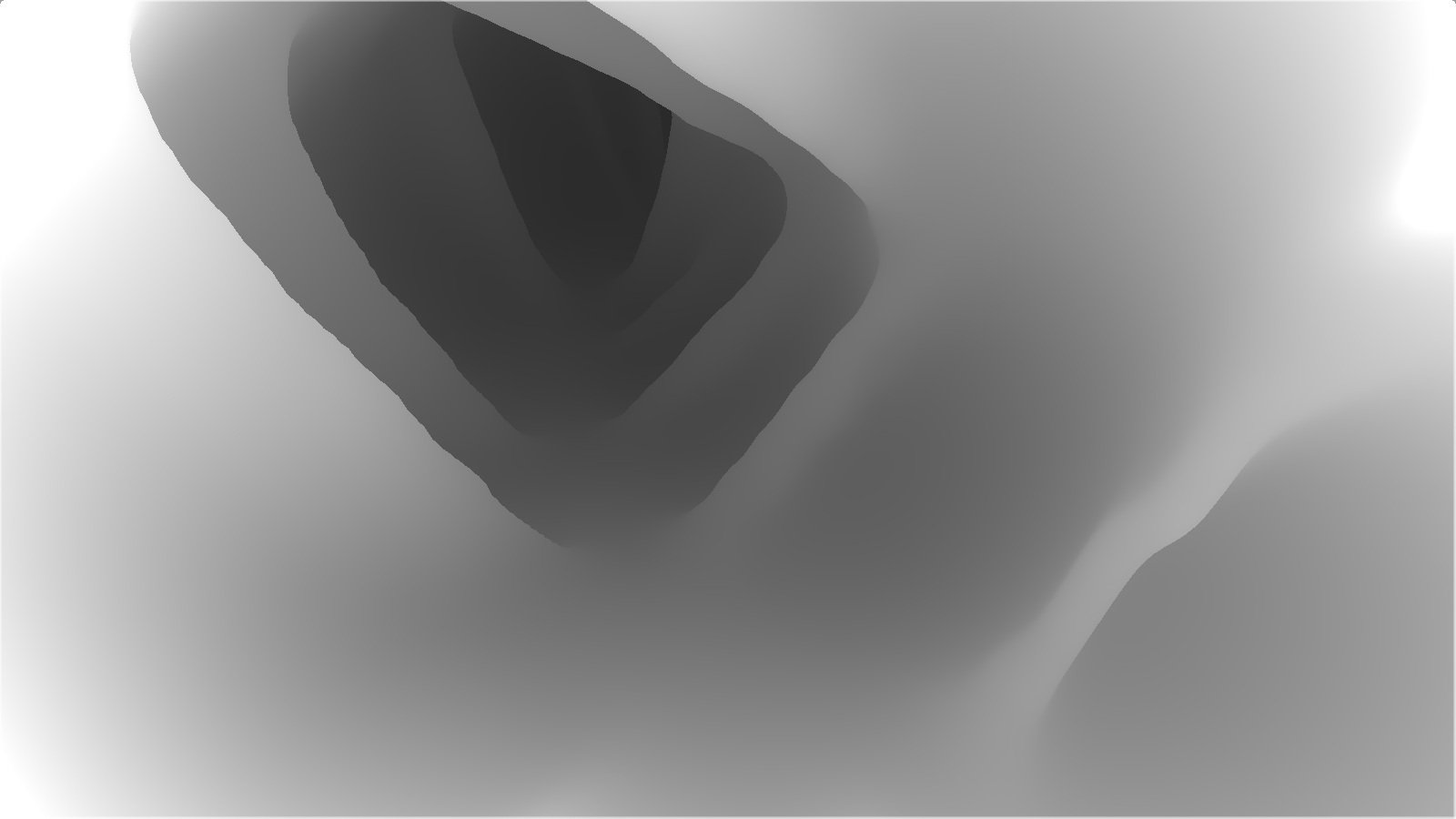}&
\includegraphics[height=0.065\textwidth]{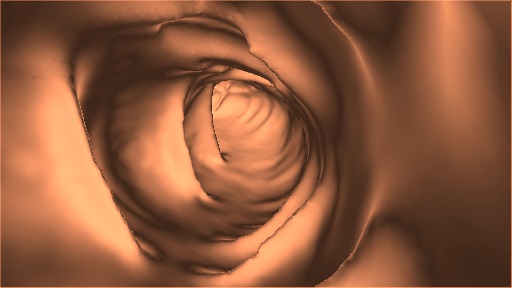}&
\includegraphics[height=0.065\textwidth]{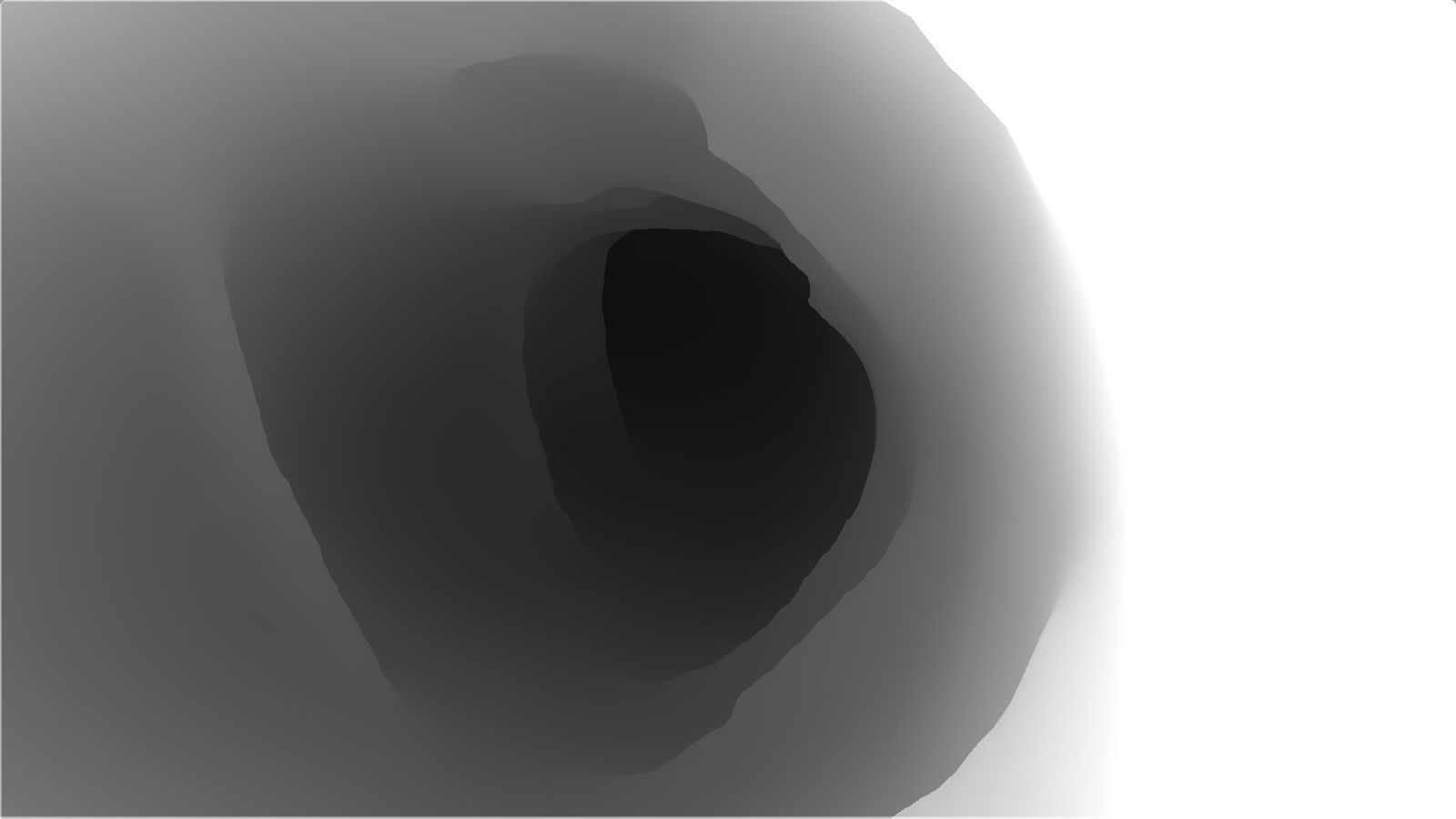}&
\includegraphics[height=0.065\textwidth]{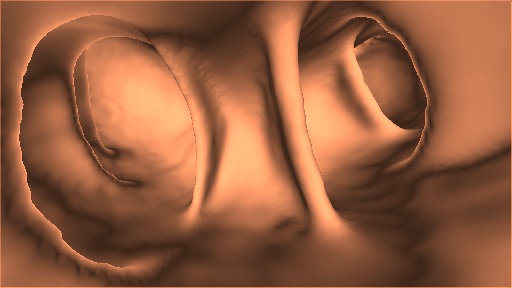}&
\includegraphics[height=0.065\textwidth]{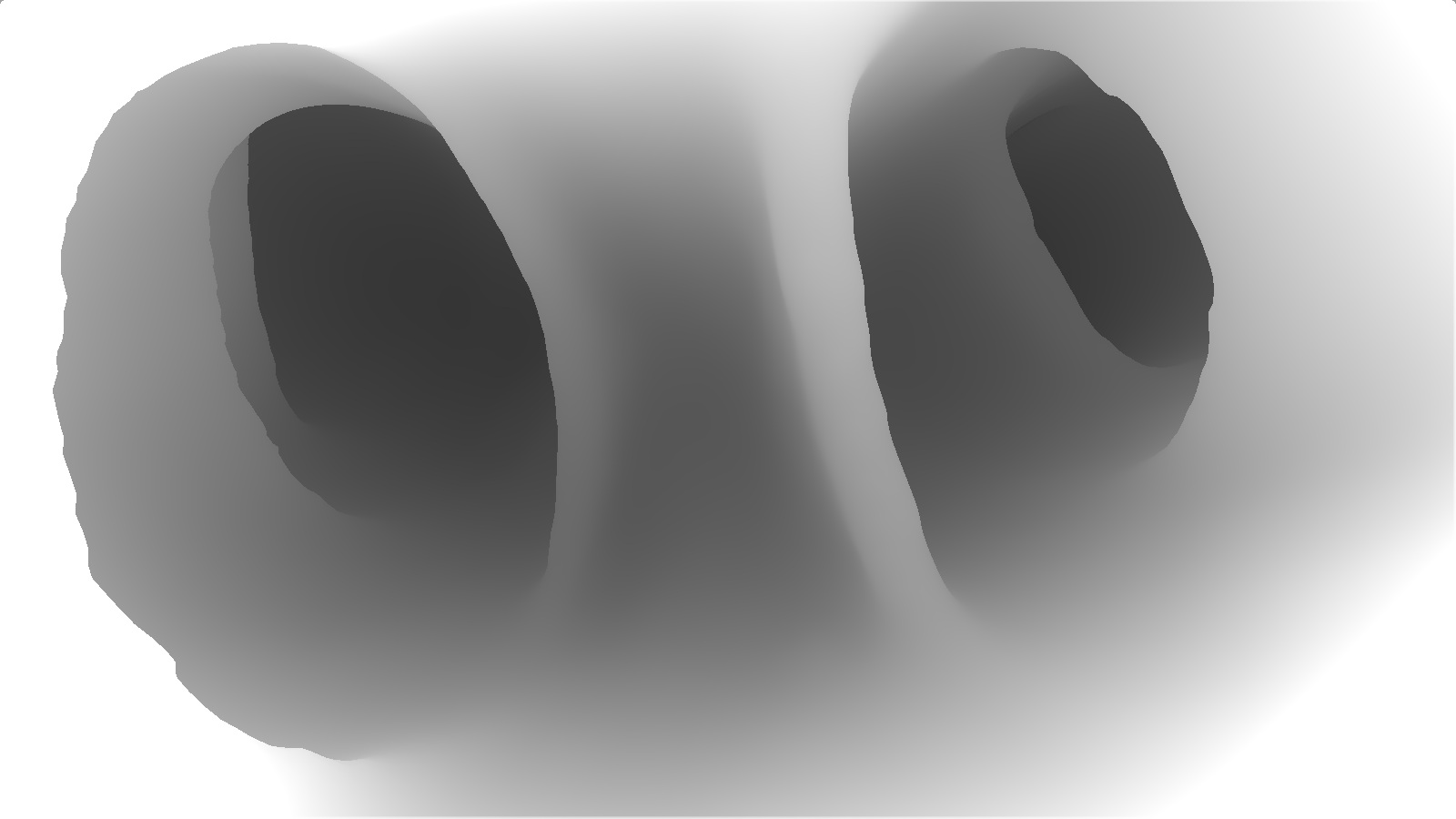}\\
\end{tabular}
\end{center}
\caption{Dictionary comprising of the RGB images and their corresponding depth maps, acquired from VC data.}
\label{fig:dictionary}
\end{figure}

We create our training RGB-Depth dictionary using 20 real VC colon datasets from the publicly available National Institute of Biomedical Imaging and Bioengineering (NIBIB) Image and Clinical Data Repository provided by the National Institute of Health (NIH). We perform electronic colon cleansing incorporating the partial volume effect \cite{wang:2006}, segmentation with topological simplification \cite{Hong:2006}, extraction of colon centerline and reconstruction of the colon surface via surface nets \cite{Gibson:1998} on the original CT images. Though the size and resolution of each CT volume varies from dataset to dataset, the general data size is approximately $512\times512\times450$ voxels and the general resolution is approximately $0.7\times0.7\times1.0$mm. In this paper, the colon surface is modeled as a topological cylinder and discretely represented by a triangular mesh.

Once we have extracted the colon surface and centerline, we simulate automated virtual fly-throughs in these 20 reconstructed 3D surface mesh models, from the cecum to the rectum, capturing the RGB images and their corresponding depth maps. These RGB images and their corresponding depth maps constitute our training dictionary. In total, we have collected 1528 RGB images and their corresponding depth maps.

The 20 VC datasets selected from the NIH repository include 15 datasets from patients with 1 or more lesions and 5 datasets with no lesions at all. This selection is random and has no relevance to any specific characteristics of the datasets.

\begin{figure}[!t]
\centering
\includegraphics[width=1\linewidth]{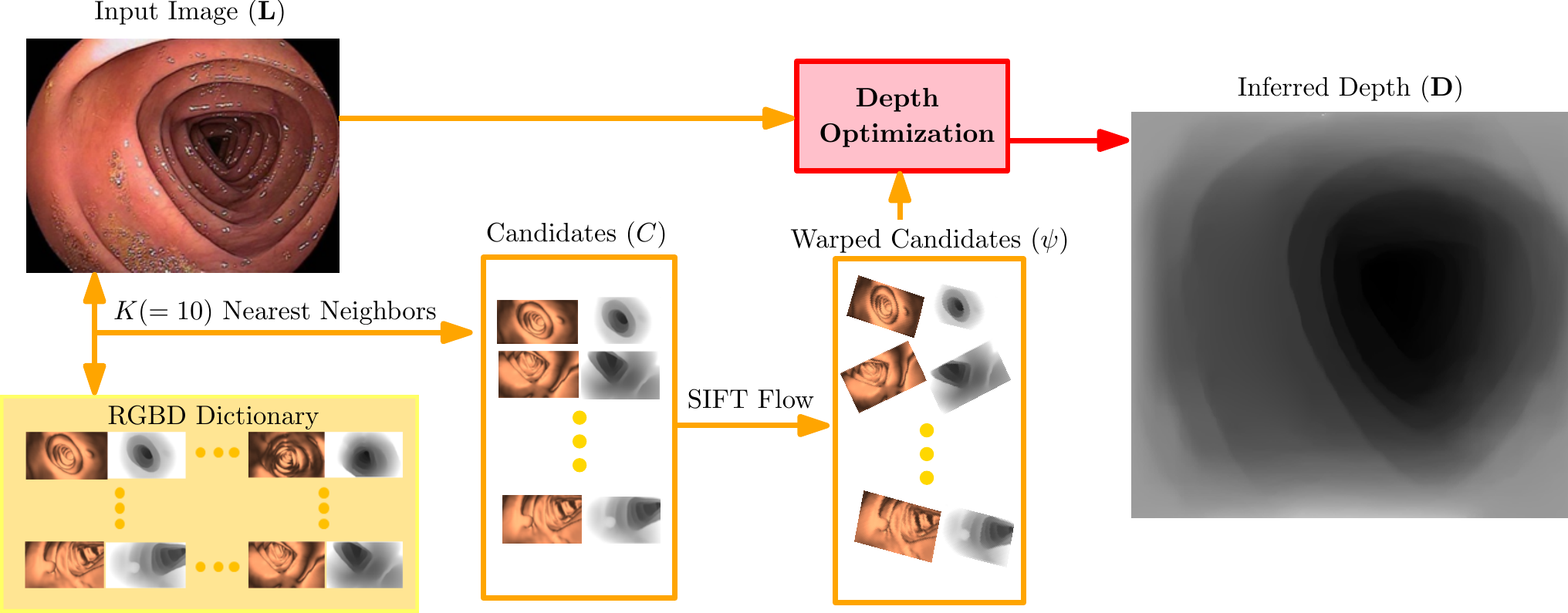}
\caption{Our pipeline for estimating depth. Given an input image, we find matching candidates in our dictionary, and warp the candidates to match the structure of the input image. We then use a global optimization procedure to interpolate the warped candidates (Eq. 1), producing per-pixel depth estimates for the input image.}
\label{fig:method}
\end{figure}

\setlength{\tabcolsep}{2pt}
\begin{figure}[!t]
\begin{center}
\begin{tabular}{cccc}
Input Frame & Candidate Images and Depths & Contribution & Resultant Depth\\
\includegraphics[width=0.23\textwidth, height=0.23\textwidth]{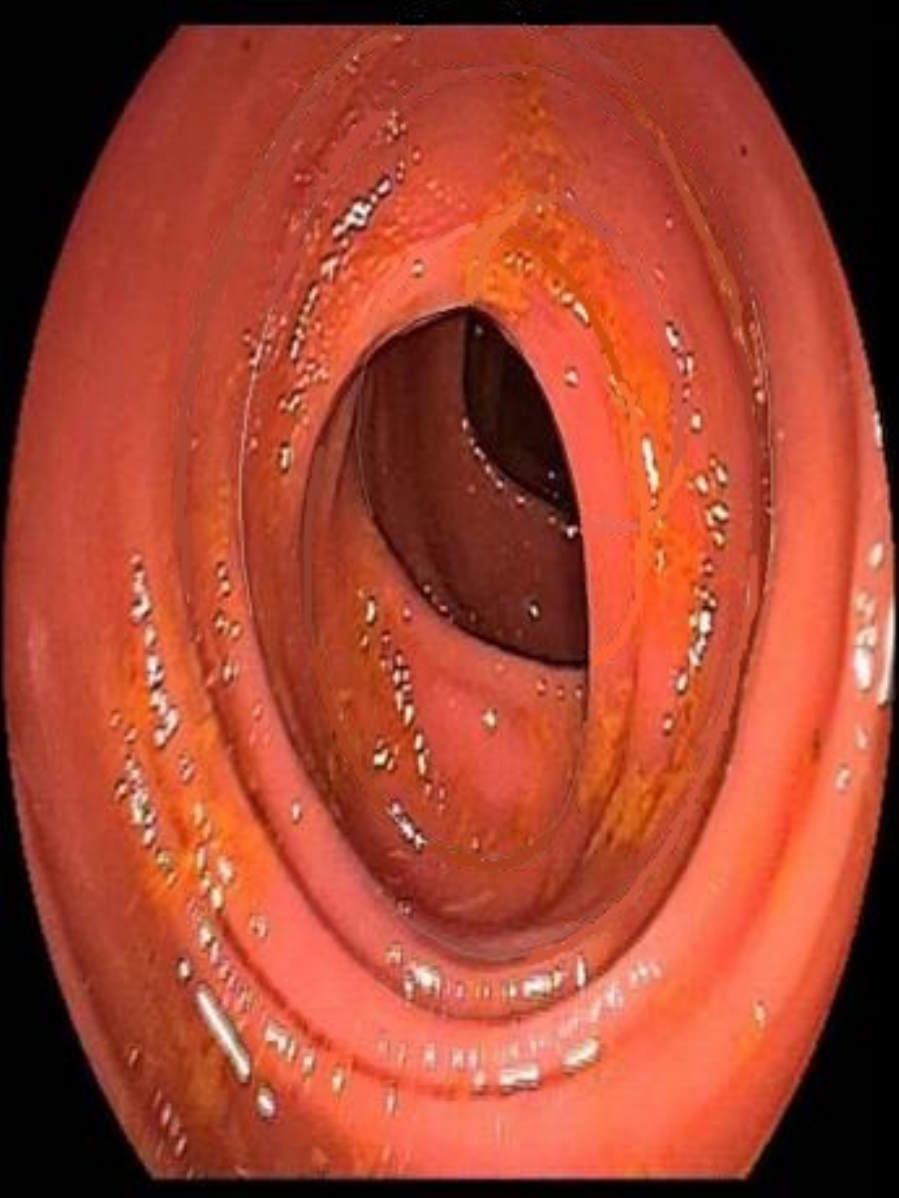}&
\includegraphics[width=0.27\textwidth, height=0.23\textwidth]{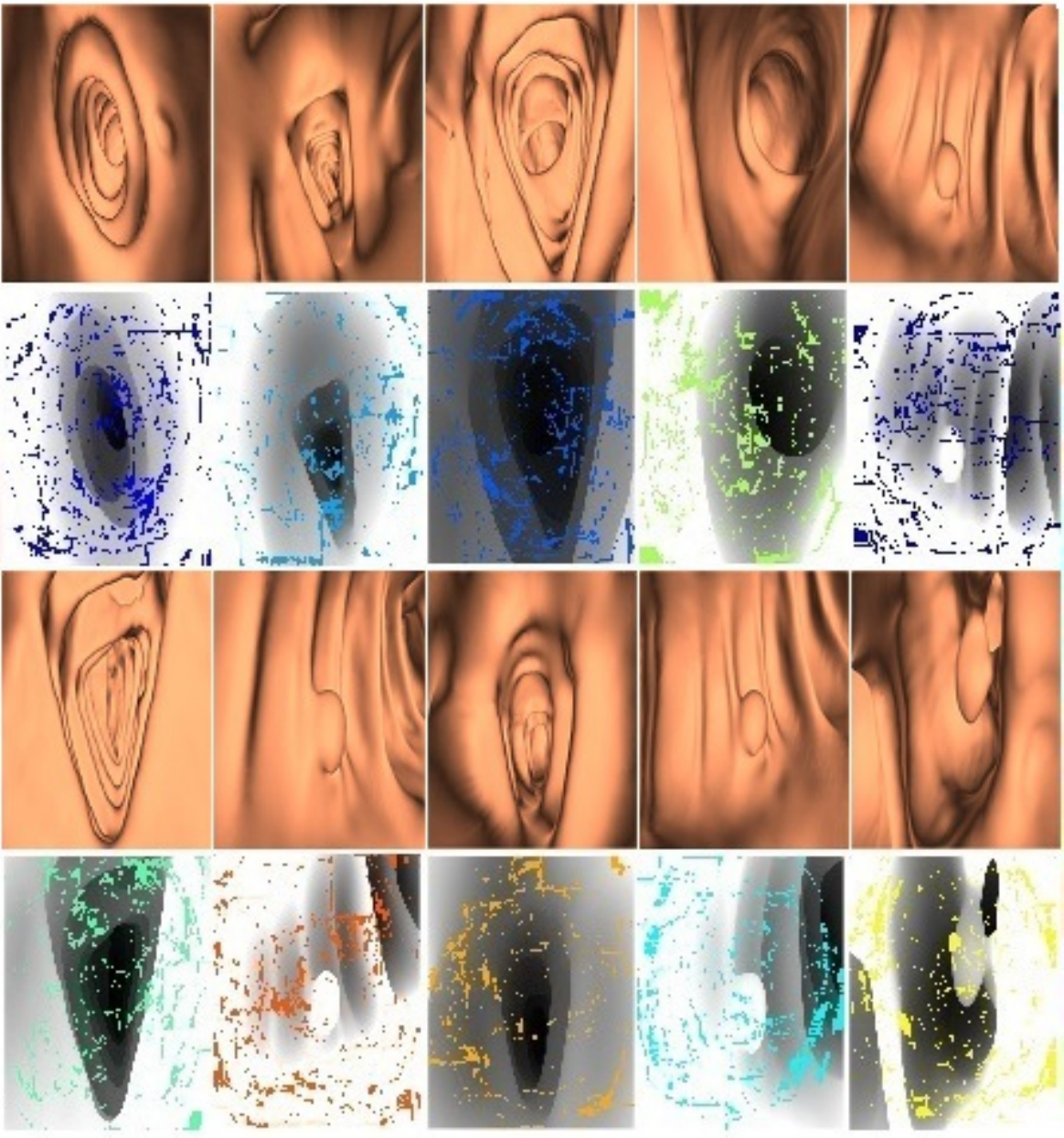}&
\includegraphics[width=0.23\textwidth, height=0.23\textwidth]{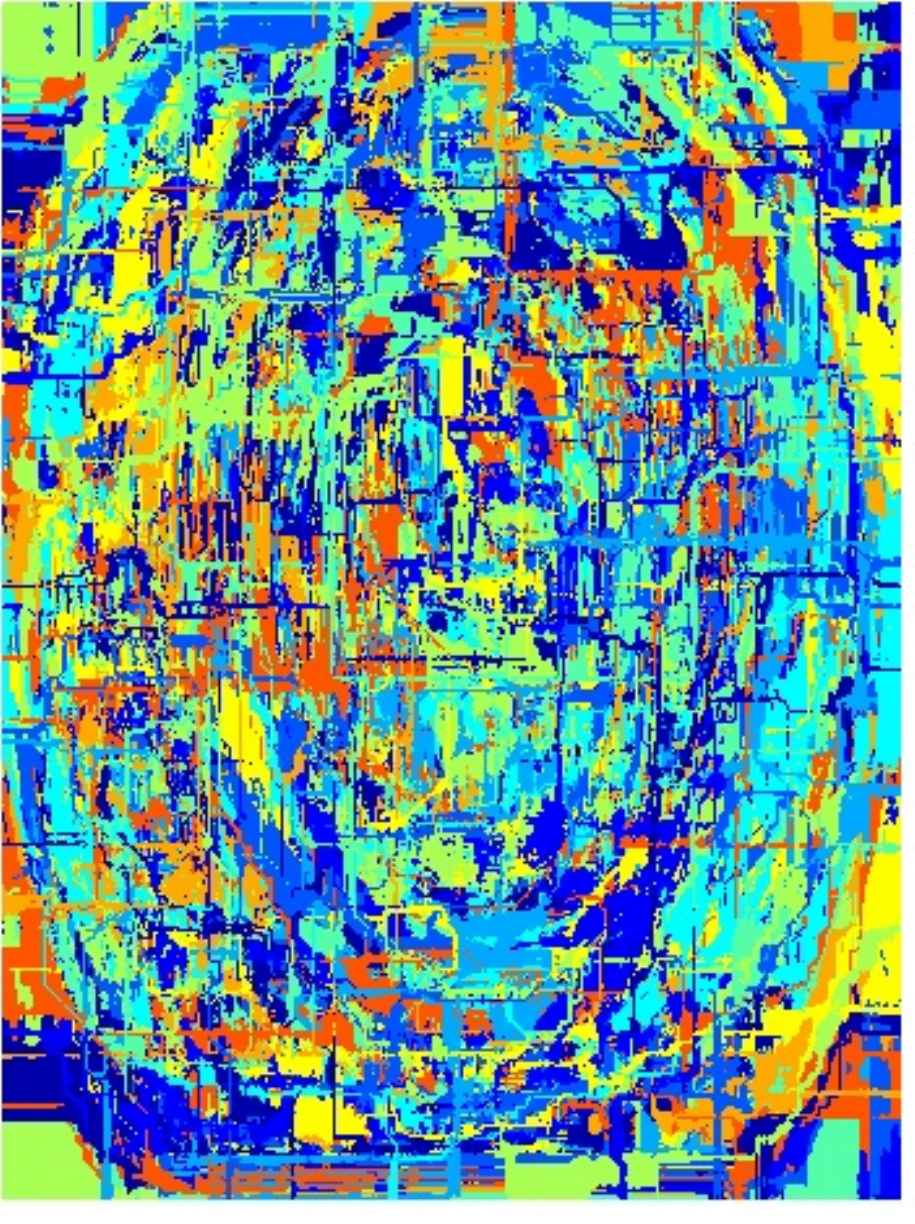}&
\includegraphics[width=0.23\textwidth, height=0.23\textwidth]{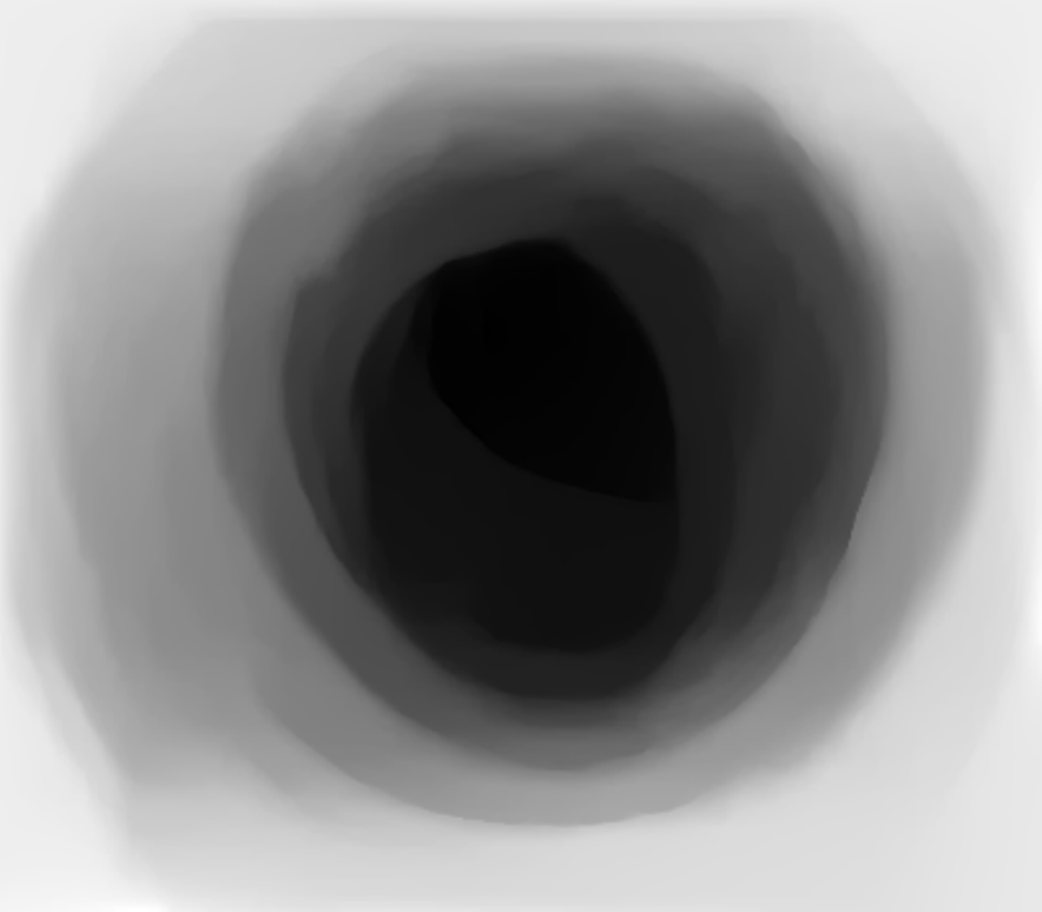}\\
(a) & (b) & (c) & (d)
\end{tabular}
\end{center}
\caption{Candidate contribution for depth estimation. Given an input image (a), we find the top 10 matching candidate images and depths (b), and infer depth for the input image (d) using our technique. The contribution image (c) is color-coded to show the sources of per-pixel depth contributed by candidate depths.}
\label{fig:depthest}
\end{figure}

\subsection{Depth Estimation}

\setlength{\tabcolsep}{4pt}
\begin{figure}[ht!]
\begin{center}
\begin{tabular}{ccc}
Input Frame & Depth & Depth--CAD\\
\includegraphics[width=0.23\textwidth, height=0.23\textwidth]{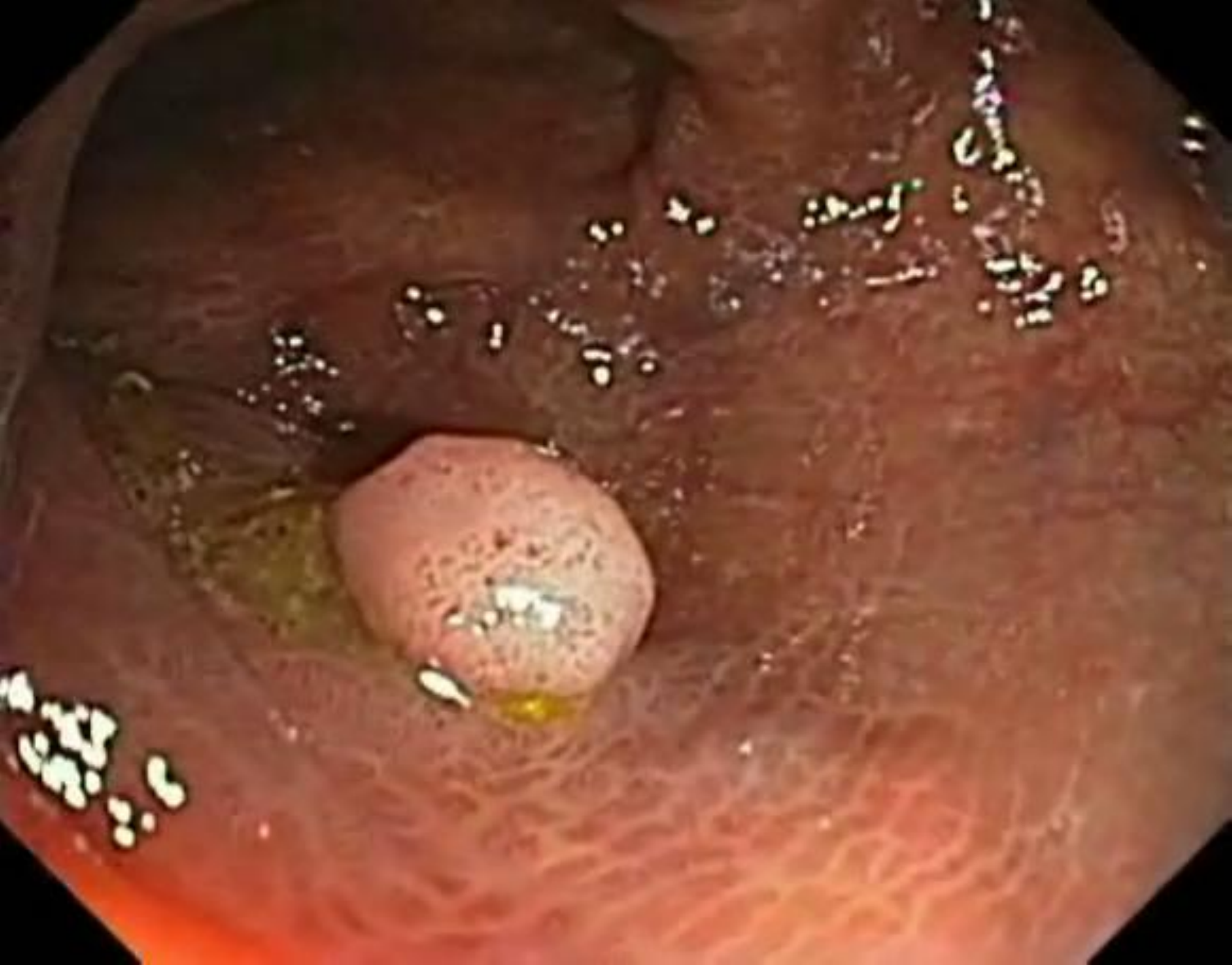}&
\includegraphics[width=0.23\textwidth, height=0.23\textwidth]{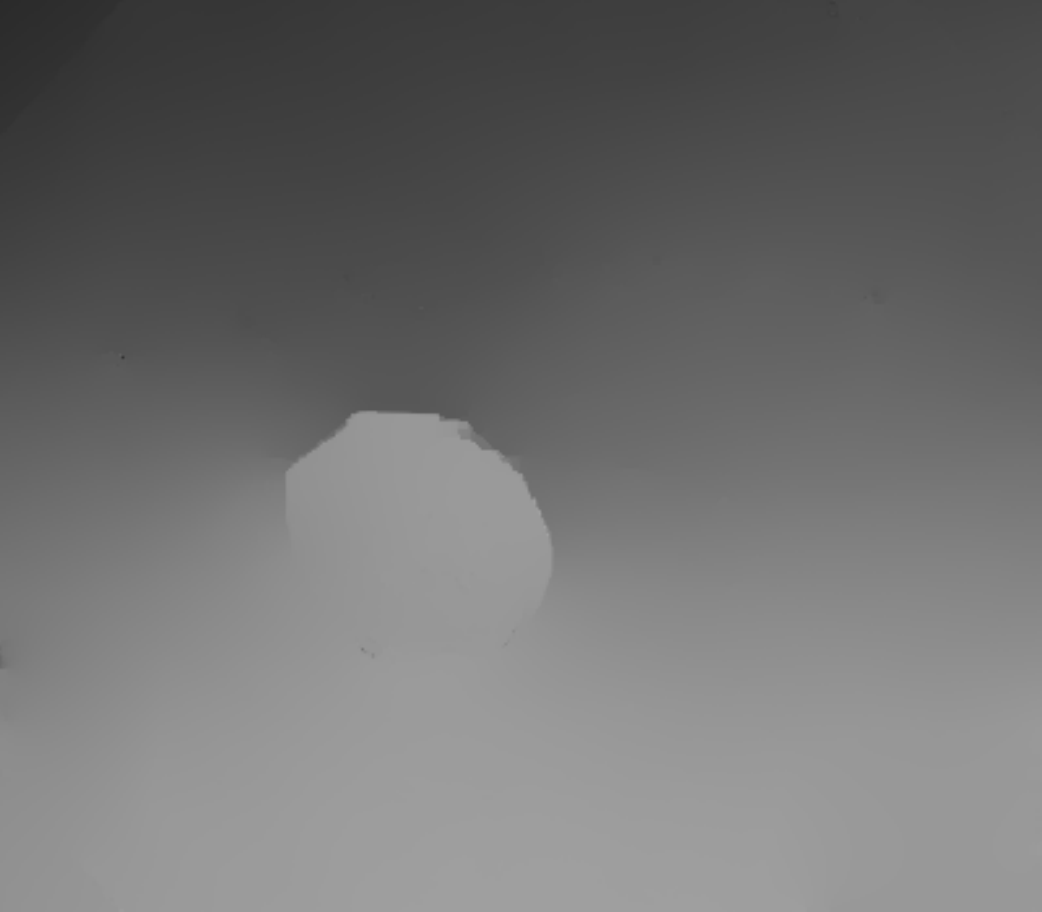}&
\includegraphics[width=0.23\textwidth, height=0.23\textwidth]{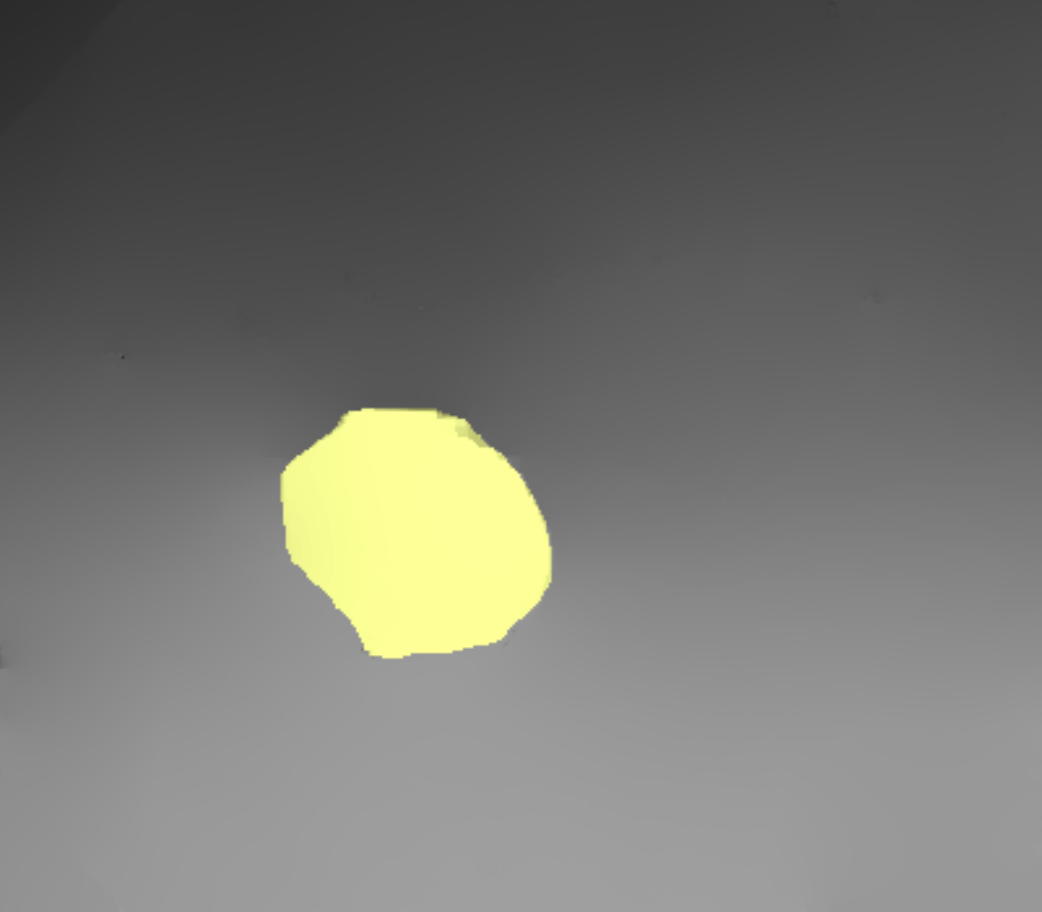}\\
\includegraphics[width=0.23\textwidth, height=0.23\textwidth]{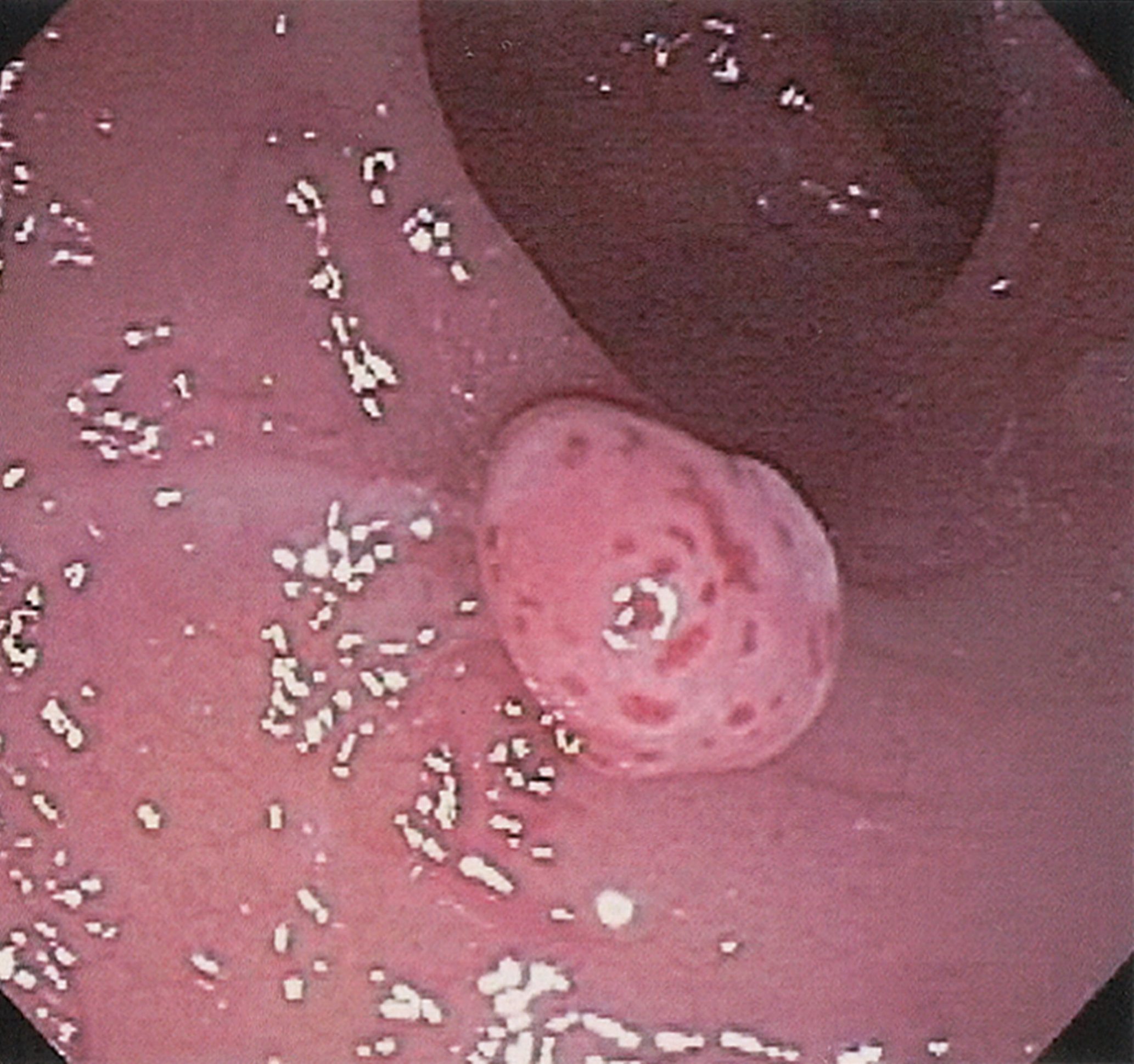}&
\includegraphics[width=0.23\textwidth, height=0.23\textwidth]{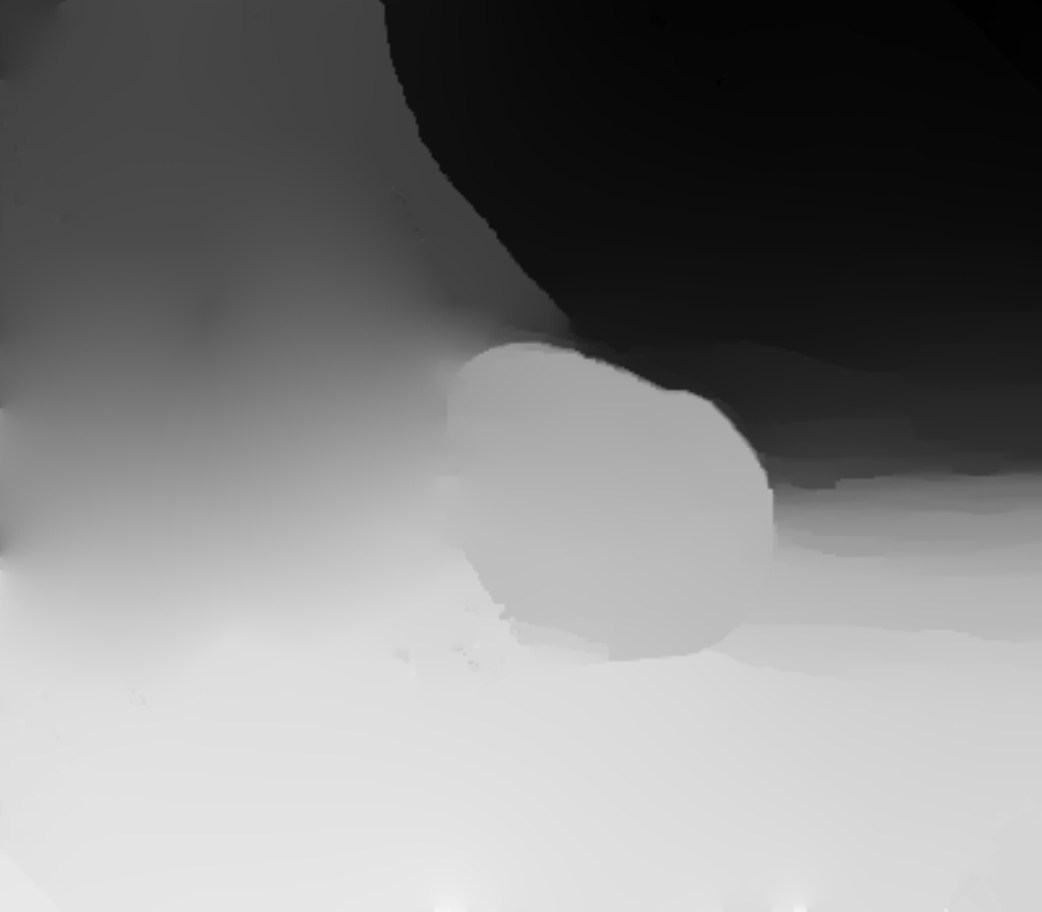}&
\includegraphics[width=0.23\textwidth, height=0.23\textwidth]{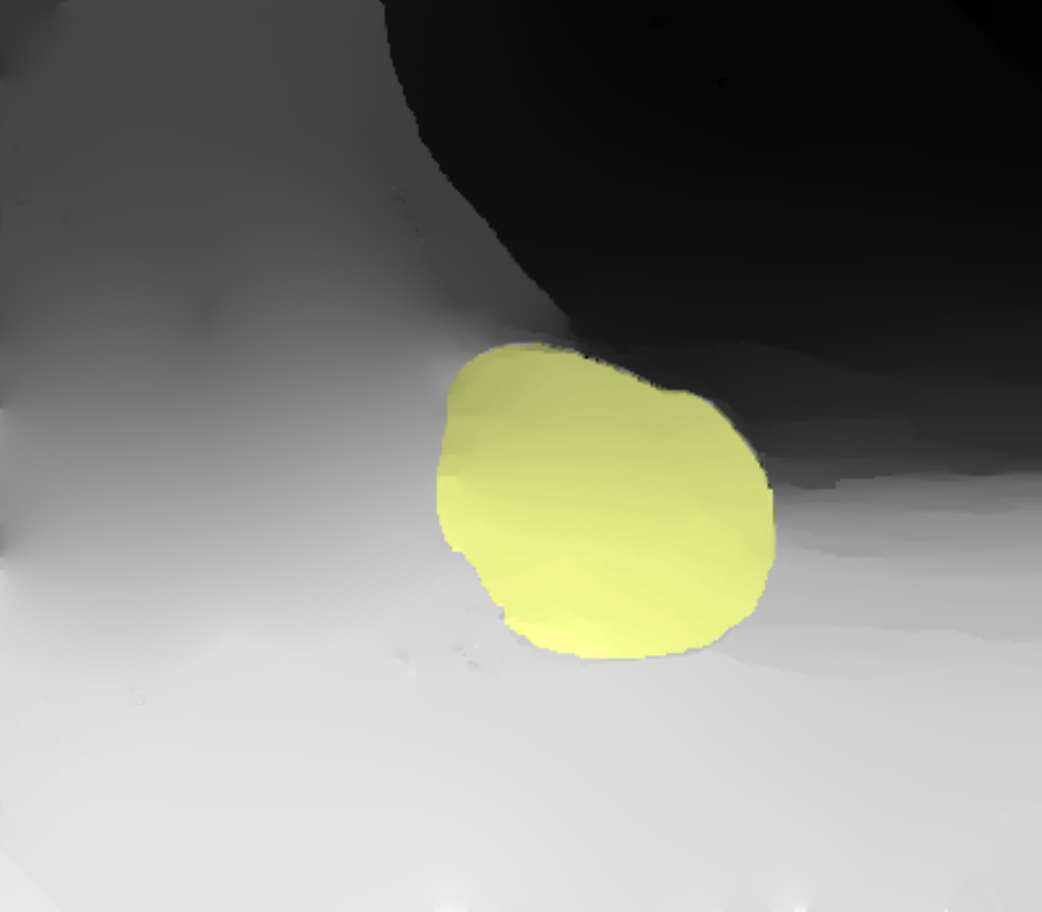}\\
\includegraphics[width=0.23\textwidth, height=0.23\textwidth]{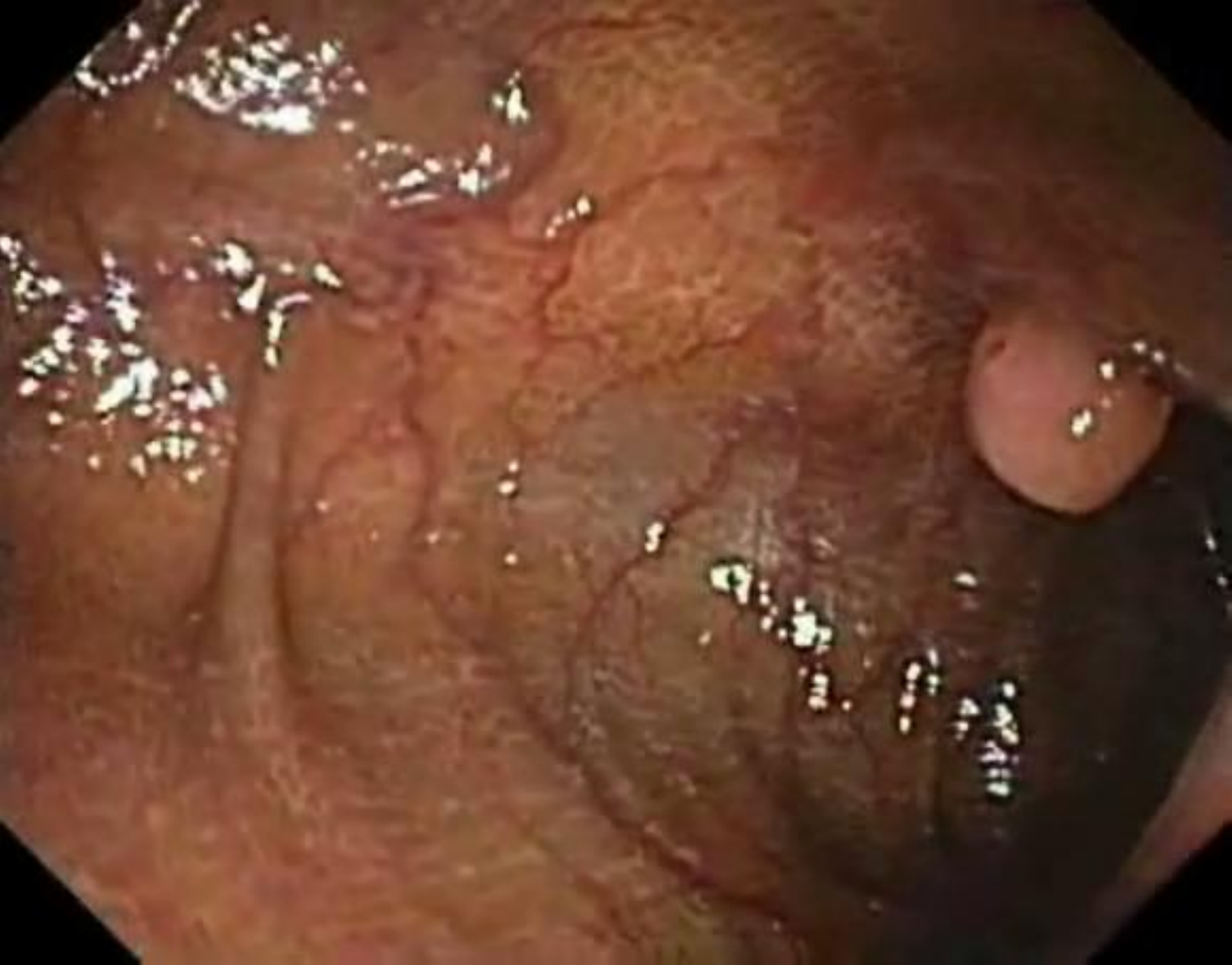}&
\includegraphics[width=0.23\textwidth, height=0.23\textwidth]{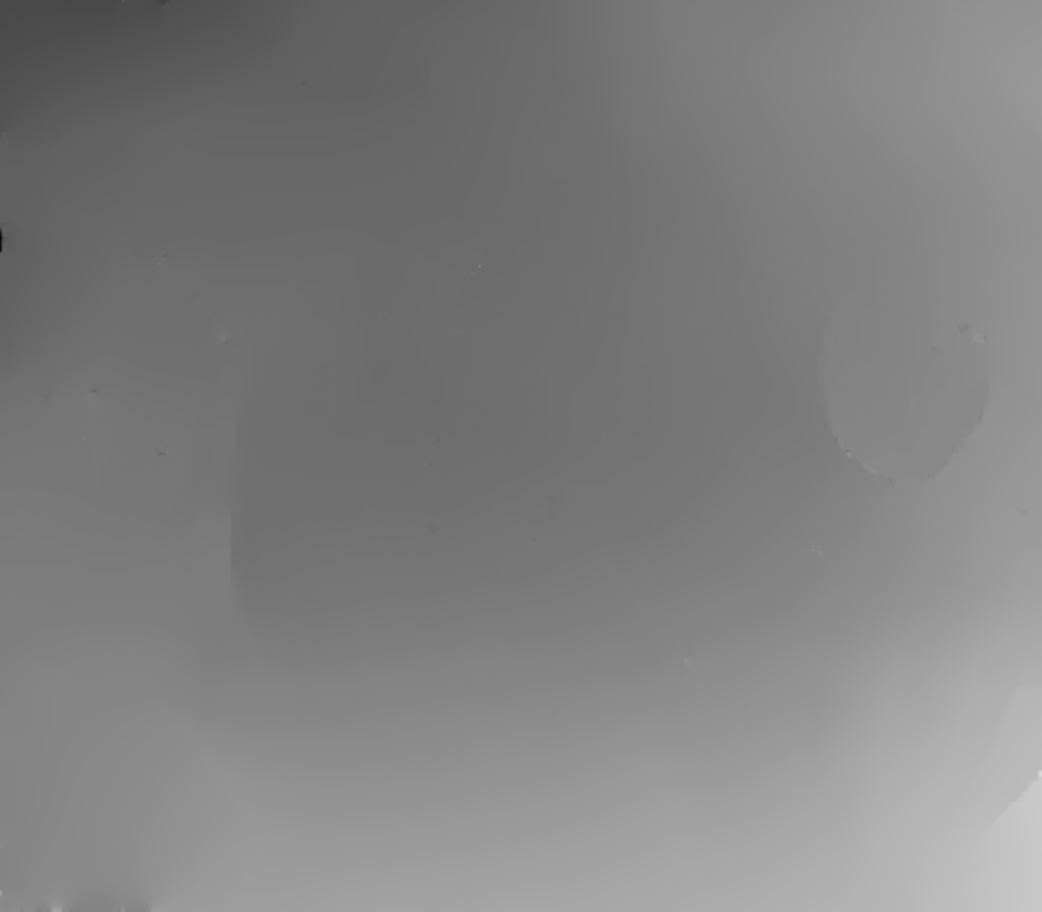}&
\includegraphics[width=0.23\textwidth, height=0.23\textwidth]{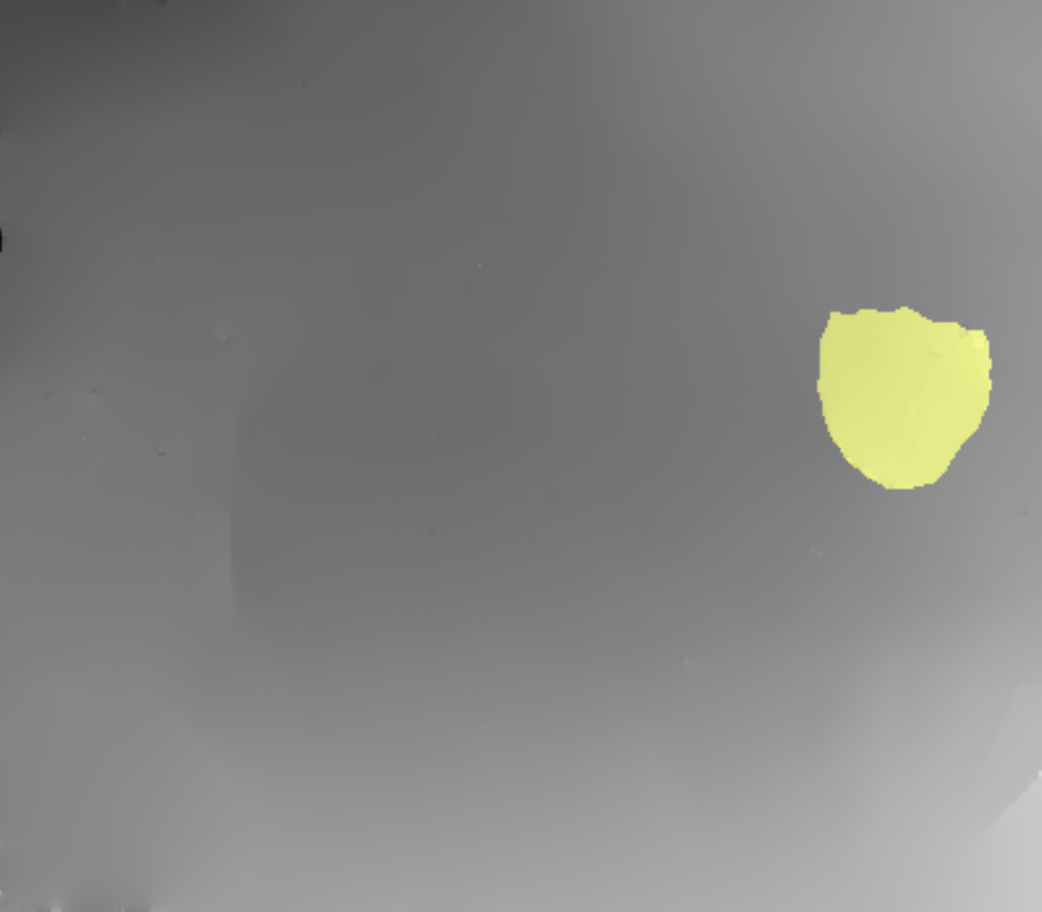}\\
\includegraphics[width=0.23\textwidth, height=0.23\textwidth]{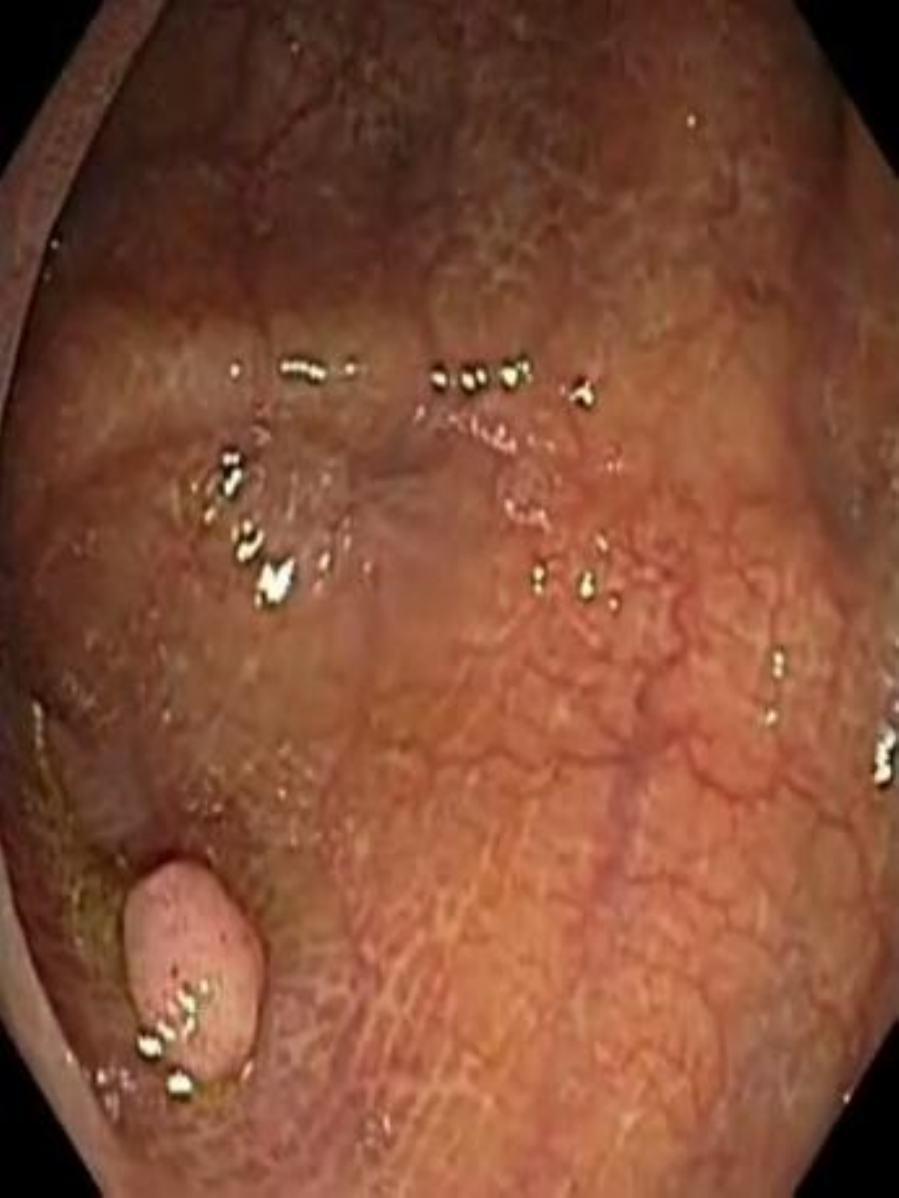}&
\includegraphics[width=0.23\textwidth, height=0.23\textwidth]{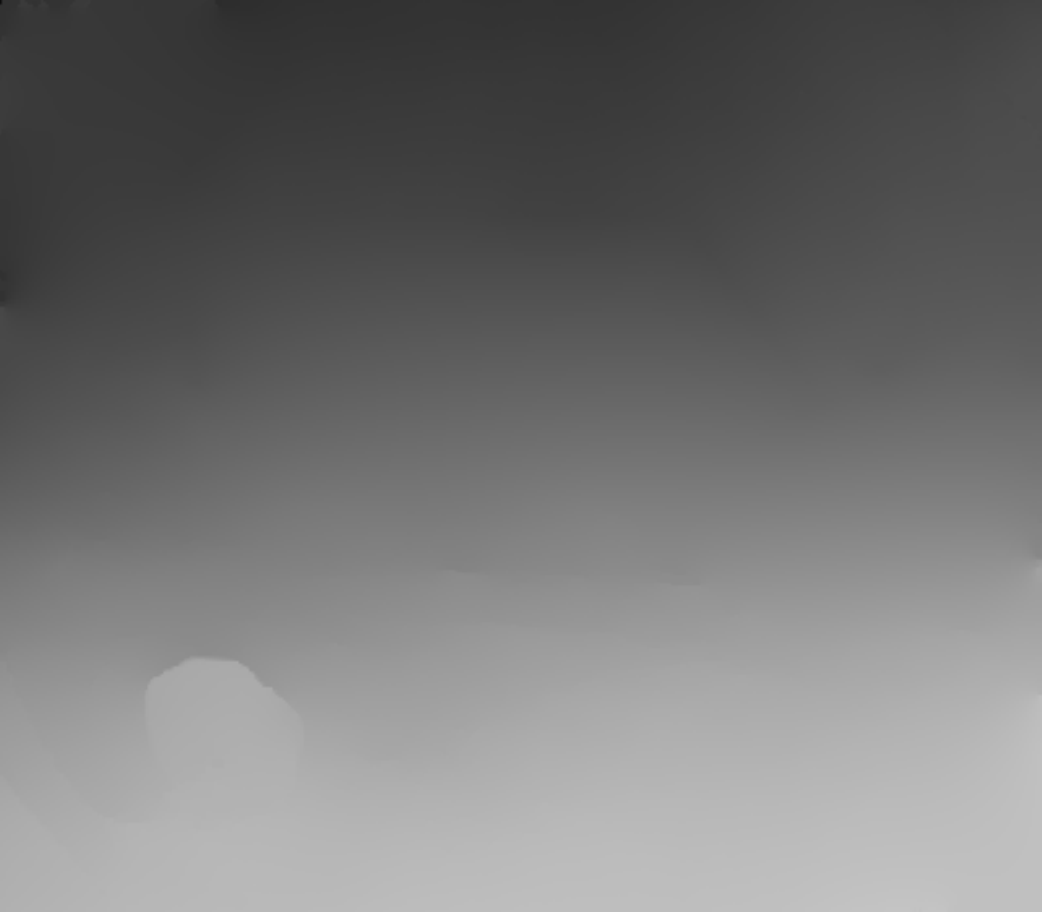}&
\includegraphics[width=0.23\textwidth, height=0.23\textwidth]{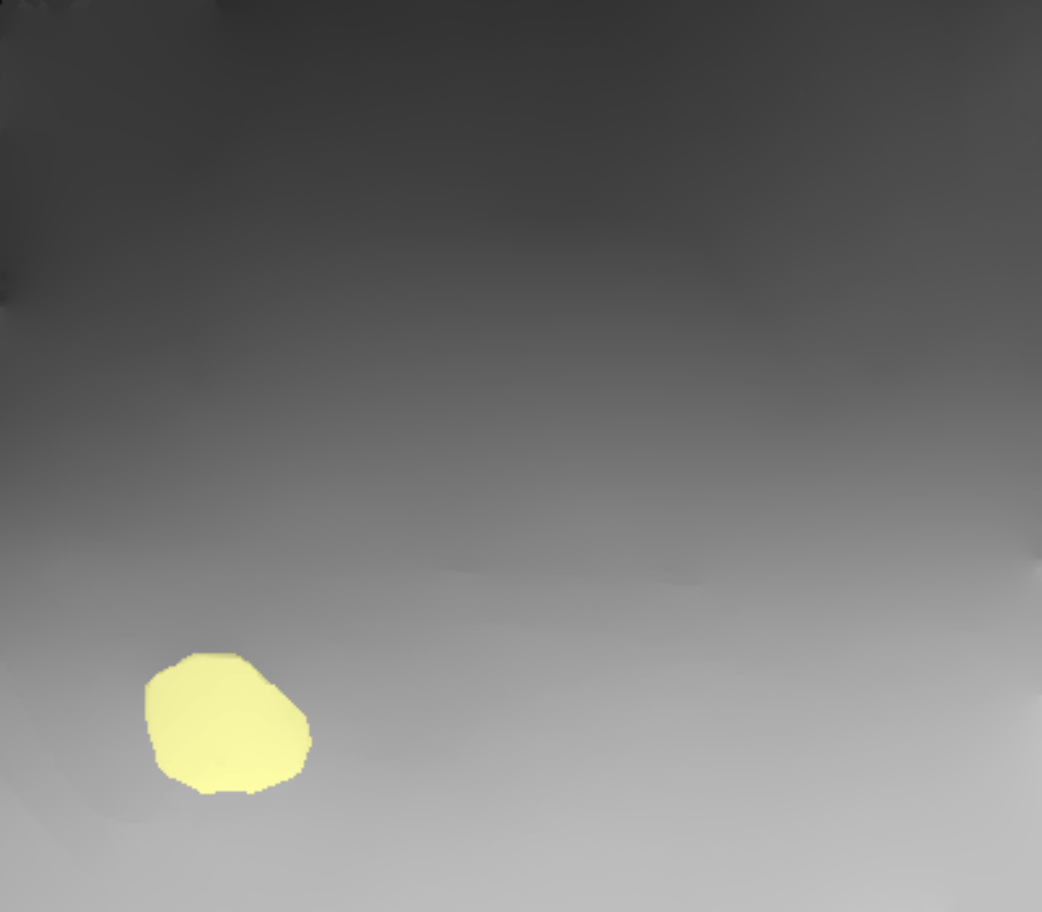}
\end{tabular}
\end{center}
\caption{Polyp visualization using our computer-aided detection algorithm.}
\label{fig:polyps}
\end{figure}

\setlength{\tabcolsep}{4pt}
\begin{figure}[ht!]
\begin{center}
\small
\begin{tabular}{cc||cc}
Input Frame & Depth Result & Input Frame & Depth Result\\
\includegraphics[width=0.235\textwidth, height=0.23\textwidth]{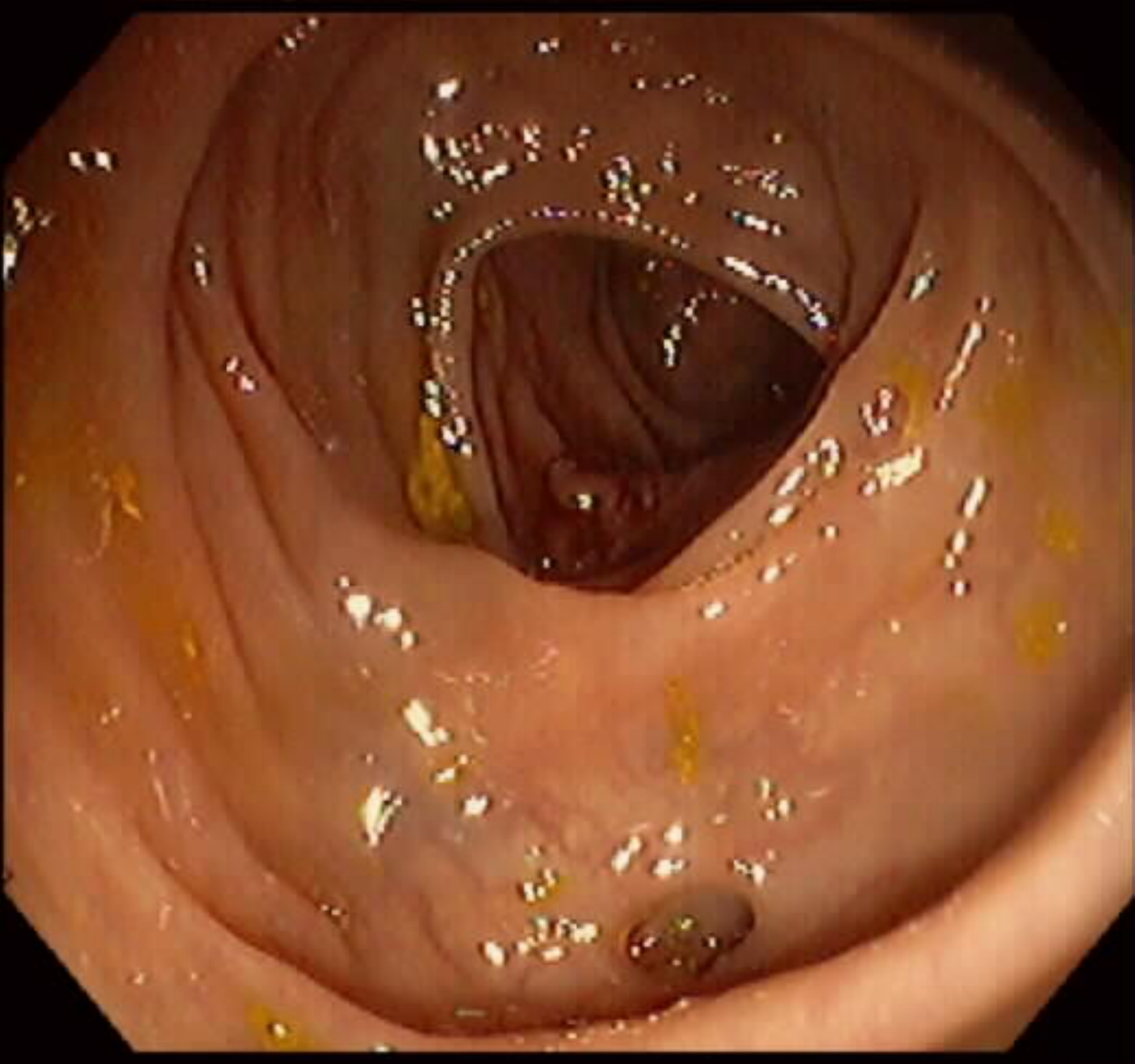}&
\includegraphics[width=0.23\textwidth, height=0.23\textwidth]{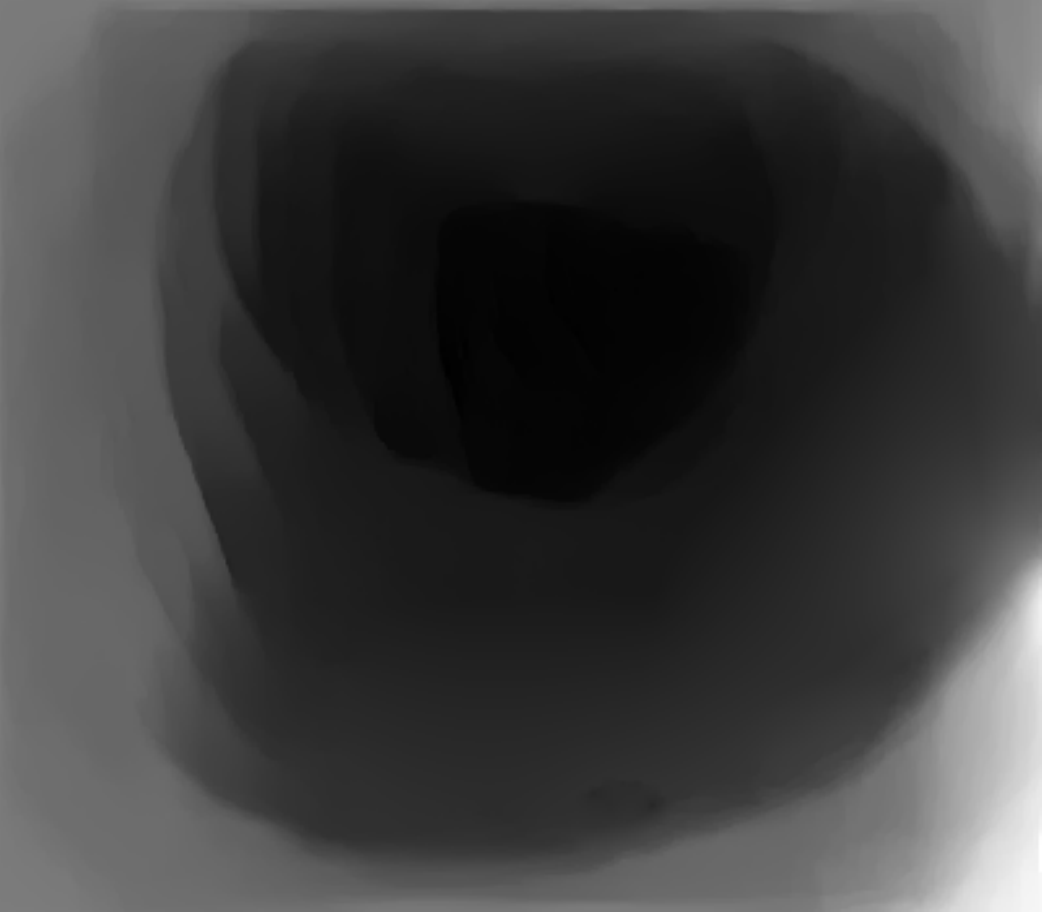}&
\includegraphics[width=0.237\textwidth, height=0.23\textwidth]{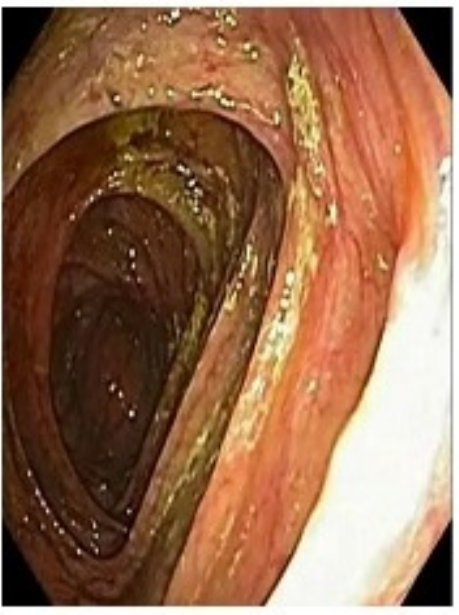}&
\includegraphics[width=0.23\textwidth, height=0.23\textwidth]{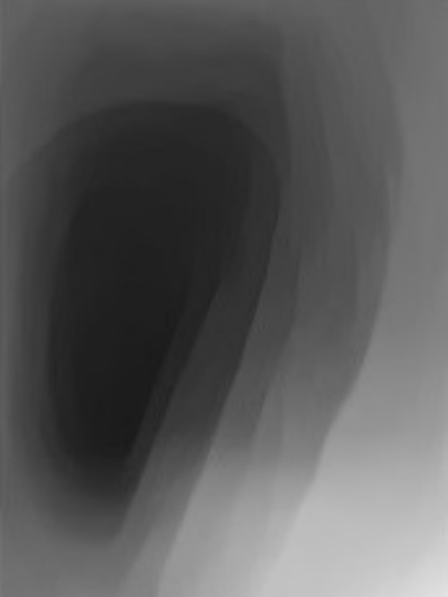}\\
\includegraphics[width=0.235\textwidth, height=0.23\textwidth]{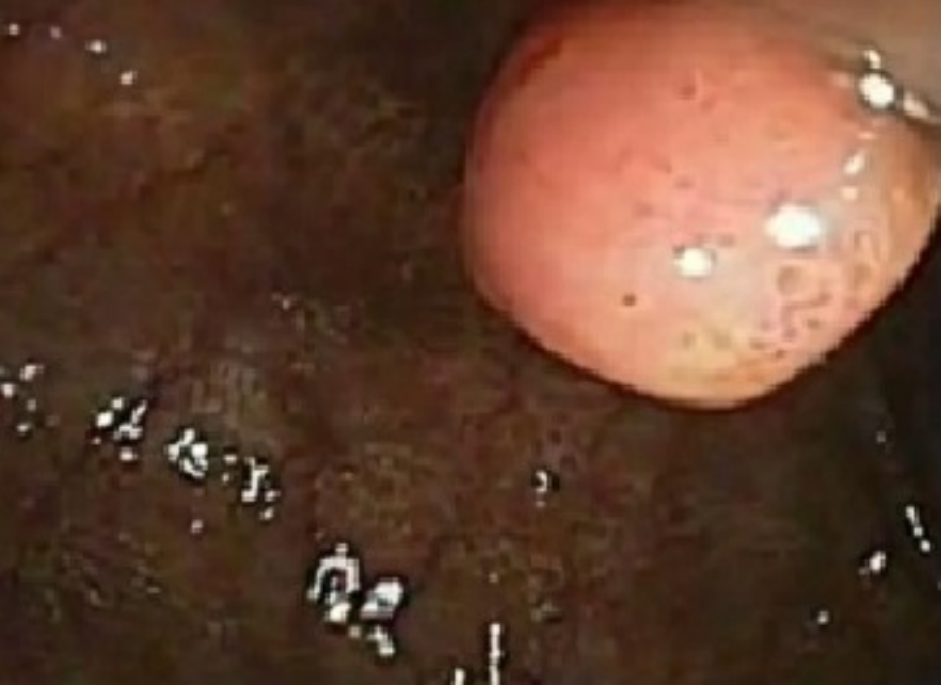}&
\includegraphics[width=0.23\textwidth, height=0.23\textwidth]{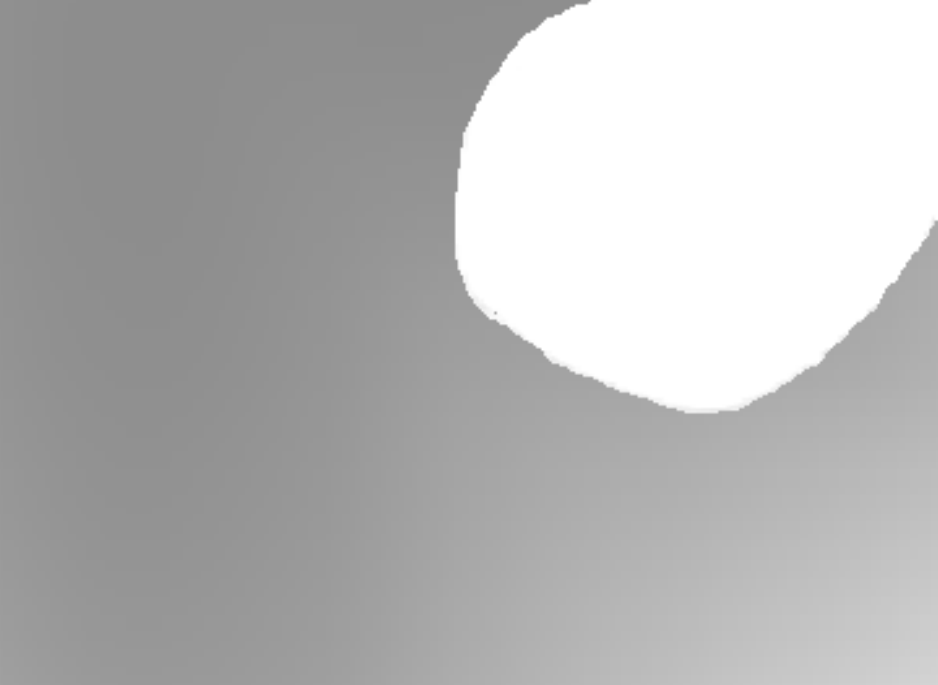}&
\includegraphics[width=0.23\textwidth, height=0.23\textwidth]{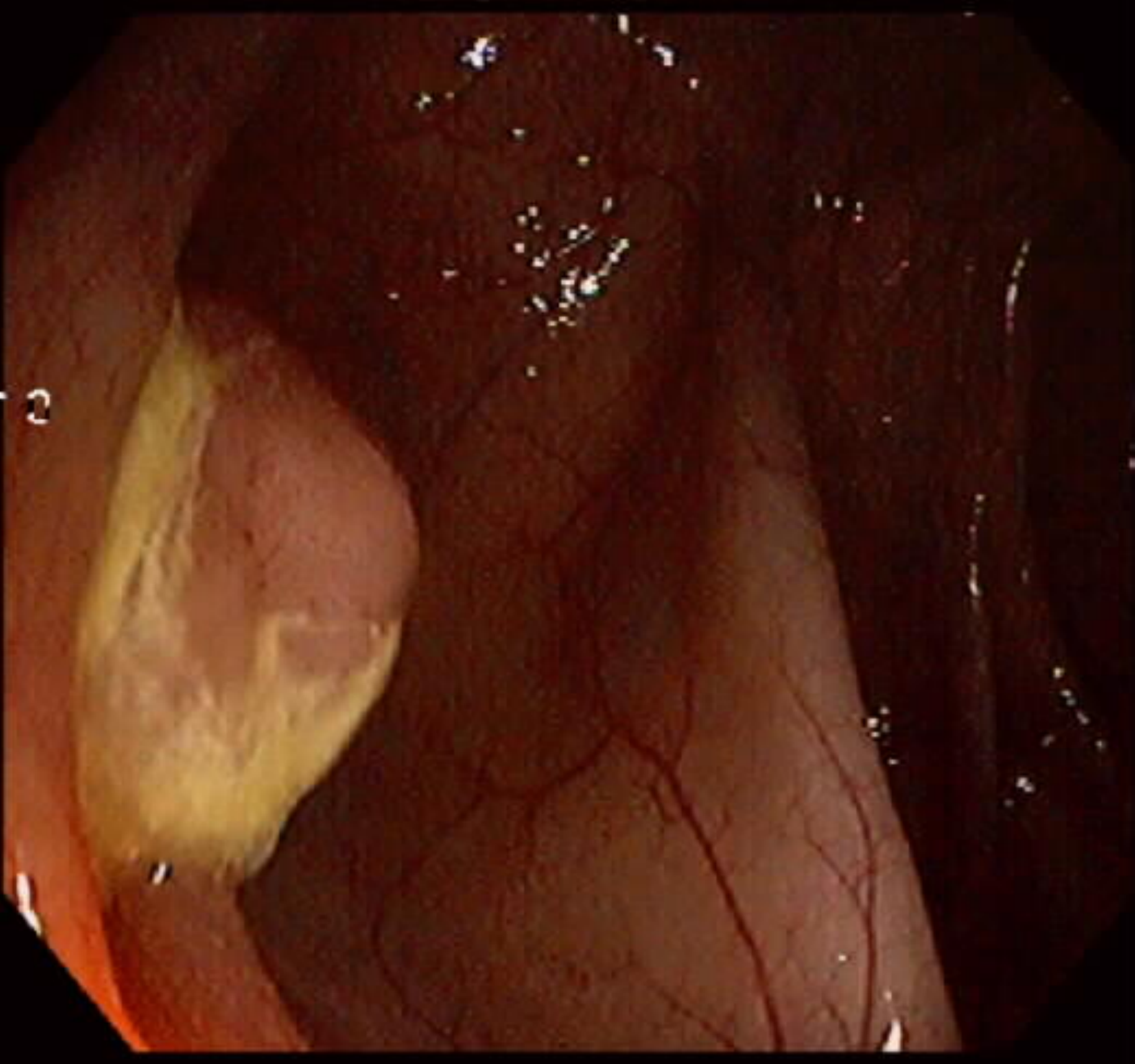}&
\includegraphics[width=0.23\textwidth, height=0.23\textwidth]{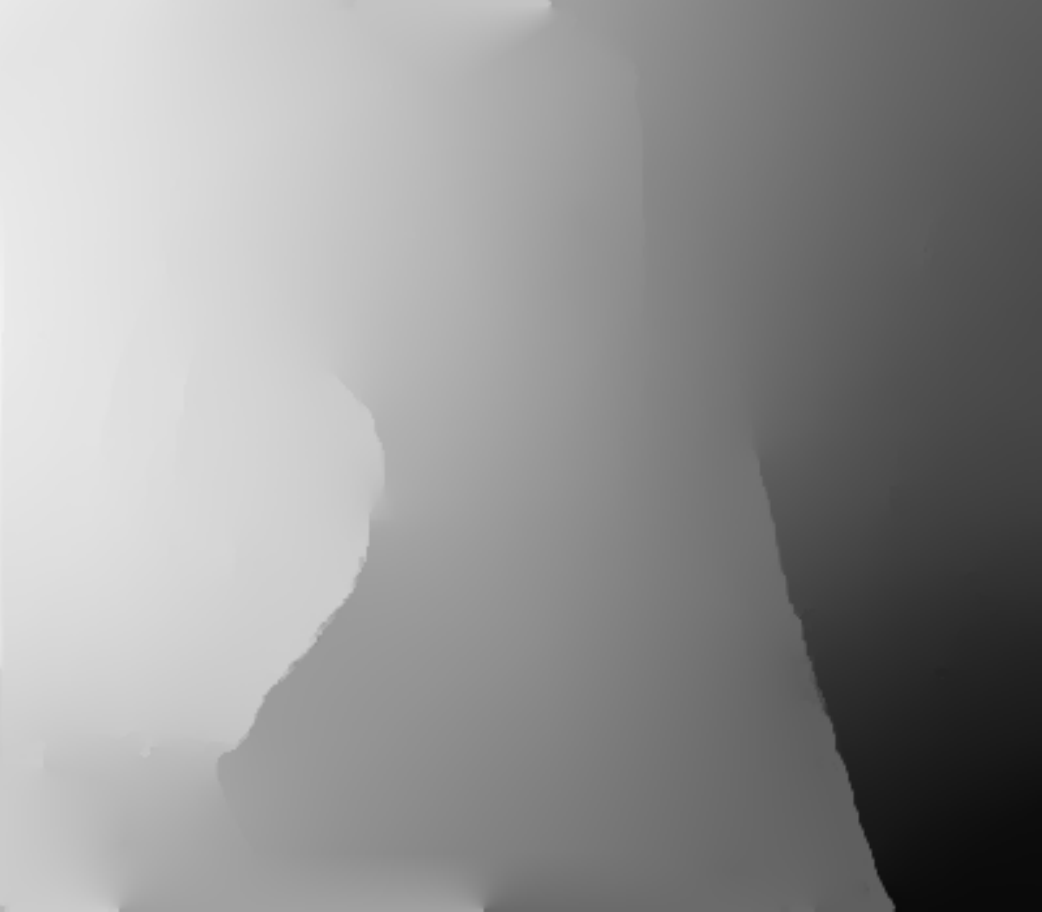}\\
\includegraphics[width=0.242\textwidth, height=0.23\textwidth]{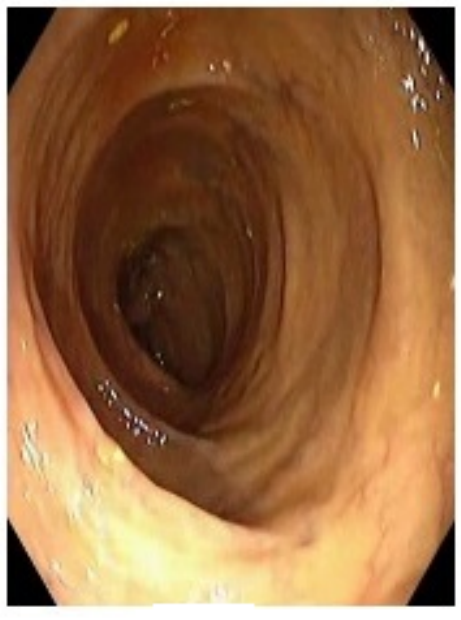}&
\includegraphics[width=0.23\textwidth, height=0.23\textwidth]{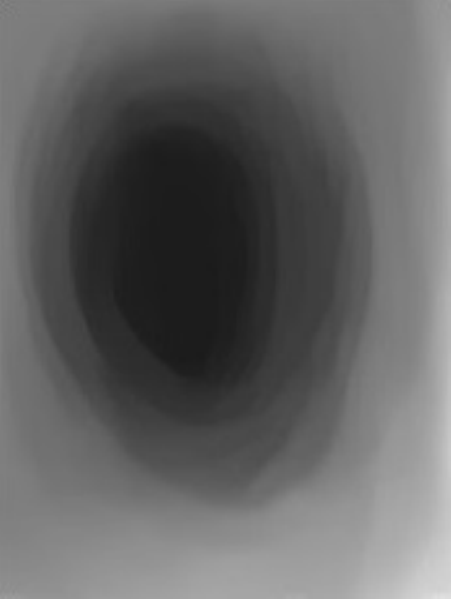}&
\includegraphics[width=0.23\textwidth, height=0.23\textwidth]{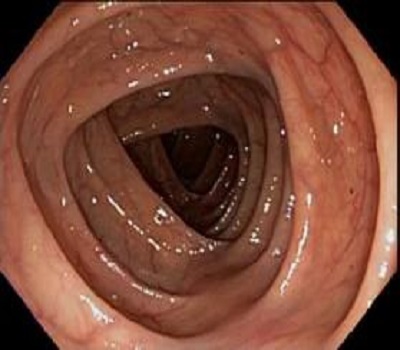}&
\includegraphics[width=0.23\textwidth, height=0.23\textwidth]{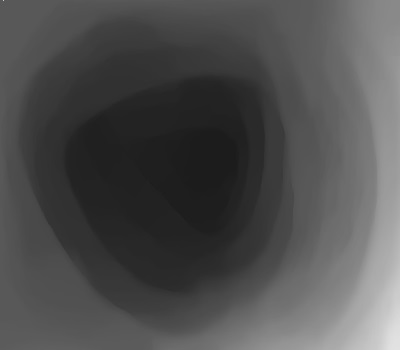}\\
\end{tabular}
\end{center}
\vspace{-2mm}
\caption{The input frames and the depth results for real colonoscopy videos.}
\label{fig:results1}
\vspace{-2mm}
\end{figure}

For the depth estimation, we have leveraged the non-parametric machine learning approach in Karsch et al. \cite{karsch:2014}. Following Karsch et al. \cite{karsch:2014}, we assume that the appearance and depth are correlated. We rely on this assumption to infer depth from a dictionary of image-depth pairs.

First, we compute gist features for the input as well as the images in our dictionary \cite{Oliva:2001,oliva:2006}. We then select the 10 nearest neighbors (with known depth maps) of our given video frame input using the k-Nearest Neighbor (kNN) algorithm, where $k=10$. The overall semantics of the candidate images and the input are deemed similar because of their proximity in the feature space. Then, we use Scale Invariant Feature Transform (SIFT) flow \cite{Liu:2011} to achieve pixel-to-pixel correspondence between candidates and the input. This results in dense scene alignment. Even though we achieved this dense alignment, candidate depths might still contain inaccuracies and are often not spatially smooth. Thus, we generate the most likely depth map representation by taking all the warped candidates into consideration using the objective function in Equation \ref{eq:depth}. The pipeline for the depth estimation is shown in Figure \ref{fig:method}, and the per pixel depth contribution of the matches from the dictionary for a given input video frame is shown in Figure \ref{fig:depthest}.

Let \textbf{L} be the input image in Figure~\ref{fig:depthest}a and \textbf{D} the depth map that we wish to infer (Figure~\ref{fig:depthest}d). We minimize a modified version of the objective function, proposed in Karsch et al.\cite{karsch:2014} :
\begin{equation}
\label{eq:depth}
-\log(P(\mathbf{D|L}))=E(\mathbf{D})=\sum_{i\in pixels} E_t (\mathbf{D}_i) + \alpha E_s(\mathbf{D}_i) + \log(Z)
\end{equation}
where $Z$ is the normalization constant of the probability, and $\alpha$ is a hyper-parameter. We select $\alpha$ to be 100. For a single image, our objective function contains two terms: data ($E_t$) and spatial smoothness ($E_s$). \emph{We take these terms as they are defined in Karsch et al.} \cite{karsch:2014}.

The data term measures the distance $\phi$, between the inferred depth map \textbf{D} and the warped candidate depths $\psi_j(C^{(j)})$:

\begin{eqnarray}
E_t(\mathbf{D}_i)&=&\sum_{j=1}^{K}\omega_{i}^{(j)}[\phi(\mathbf{D}_i-\psi_j(C_i^{(j)})) + \\
&&[\phi(\nabla_{x}\mathbf{D}_{i}-\psi_{j}(\nabla_{x}C_{i}^{(j)})) + \phi(\nabla_{y}\mathbf{D}_{i}-\psi{j}(\nabla_{y}C_{i}^{(j)}))]], \nonumber
\end{eqnarray}
where $\omega_{i}^{(j)}$ is a confidence measure of the accuracy of the $j^{th}$ match warped depth at pixel $i$, and $K$ is the total number of matches (Figure~\ref{fig:depthest}b, $K=10$). We compute this confidence measure by comparing per-pixel SIFT descriptors, obtained during the SIFT flow computation, of both the input image and the candidate images:
\begin{equation}
\omega_{i}^{(j)} = (1 + e^{||\mathbf{S}_i - \psi_j (S_{i}^{j})||-0.5/0.01})^{-1}
\end{equation}
where $\mathbf{S}_i$ and $S_{i}^{j}$ are the SIFT feature vectors at pixel $i$ in candidate image $j$. Notice that the candidate image SIFT features are computed first, and then warped using the warping function ($\psi_j$) calculated with SIFT flow. Apart from absolute depth, depth gradients along $x$ and $y$ $(\nabla_x,\nabla_y)$ are also taken into account and similarity is enforced among candidate depth gradients and inferred depth gradients.

The spatial smoothness term encourages smoothness especially in regions where the input image has small intensity gradients, that is, in specular highlights and veins:
\begin{equation}
E_s(\mathbf{D}_i)=s_{x,i}\phi(\nabla_{x}\mathbf{D}_i) + s_{y,i}\phi(\nabla_y\mathbf{D}_i).
\end{equation}
The depth gradients along $x$ and $y$ ($\nabla_{x}\mathbf{D}$, $\nabla_{y}\mathbf{D}$) are modulated by soft thresholds (sigmoidal functions) of frame gradients in the same directions ($\nabla_{x}\mathbf{L}$, $\nabla_{y}\mathbf{L}$), namely $s_{x,i}=(1+e^{(\|\nabla_x\mathbf{L}_i\|-0.05)/0.01})^{-1}$ and $s_{y,i}=(1+e^{(\|\nabla_y\mathbf{L}_i\|-0.05)/0.01})^{-1}$.

We use iteratively reweighted least squares (IRLS) \cite{LiuThesis:2009} to minimize our objective function. IRLS approximates the objective by a linear function of the parameters, and solves the system by minimizing the squared residual, repeating until convergence.

Darker pixels indicate farther position and lighter pixels indicate closer position to the camera. All depth maps are displayed in the same log scale.

\subsection{Computer-Aided Detection of Polyps}
We use depth gradient profiling to detect polyps from our dense depth map representation. This is done by scanning the depth map with a $48\times48$ pixel window and marking the pixels with lighter pixel values, if a sharp increase in depth gradient is detected in the window. We follow the window with this gradient above and below and see if this forms an arc, encapsulating the lighter pixels. If the arc is found, we propagate the lighter pixel values within the depth threshold until the depth gradient boundary is reached. If the arc is not found, we mark these pixels as seen and repeat the process. If arcs have been seen, based on the depth gradient profiling, we annotate the frame as containing a polyp; otherwise we ignore the frame.

\subsection{Implementation and Parameter Settings}
The depth representation and the computer-aided detection modules were initially implemented in Matlab but then converted to C code for increased performance (the increase in performance was at least 4 fold). In order to compute the dense depth map representation for the whole video, we compute in parallel 100 frames at a time. An estimated time for computing the representations for 20,000 frames is 12--15 minutes. For our depth representation, we precompute the gist features for our dictionary such that these do not have to be computed over and over again for future runs.

The parameter settings for the depth representation module are explained as follows. (1) $K=10$: The results become stable at this point and do not improve beyond it. Increasing beyond this value also causes significant loss of performance without any gain in accuracy. (2) $\alpha=100$: This setting is the optimal smoothing tradeoff that we can achieve while preserving the essential features. We have verified this value with medical experts by showing them multiple results with different $\alpha$ values and they picked this value as the most suitable for our purposes.

\section{EVALUATION}
\label{sec:evaluation}

We evaluate our framework by performing cross validation on our created dictionary, presenting the results with and without distortion, comparing against a representation approach, and evaluating our CAD algorithm on our manually created dataset.

\subsection{Dictionary Evaluation}
The evaluation was performed on an Intel Core i7-3720QM 2.60GHz machine with 4GB of RAM. We present our evaluation of the dictionary and the results for the real endoscopic images with respect to this machine configuration.

\setlength{\tabcolsep}{0.9pt}
\begin{figure}[ht!]
\begin{center}
\begin{tabular}{ccc}
Input Frame & Ground Truth & Depth Result\\
\includegraphics[width=0.23\textwidth, height=0.23\textwidth]{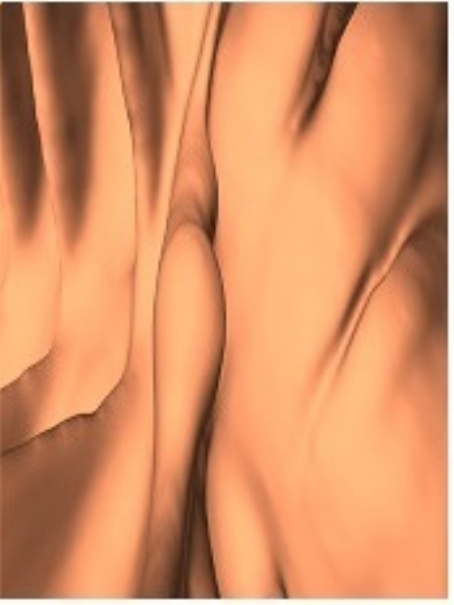}&
\includegraphics[width=0.23\textwidth, height=0.23\textwidth]{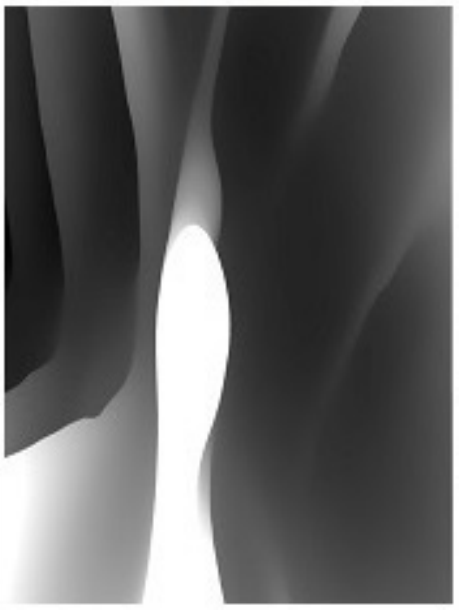}&
\includegraphics[width=0.23\textwidth, height=0.23\textwidth]{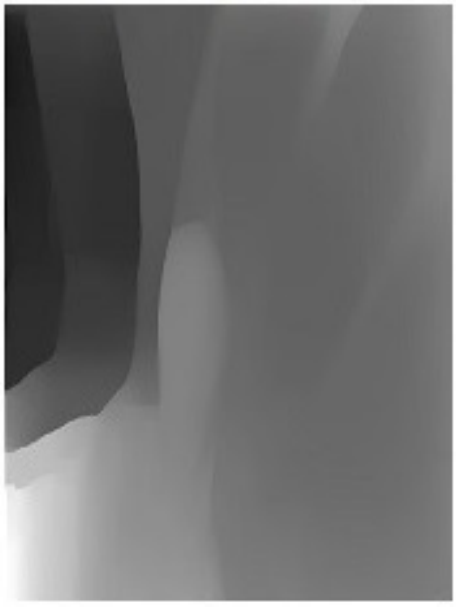}\\
(a) & (b) & (c)\\
\includegraphics[width=0.23\textwidth, height=0.23\textwidth]{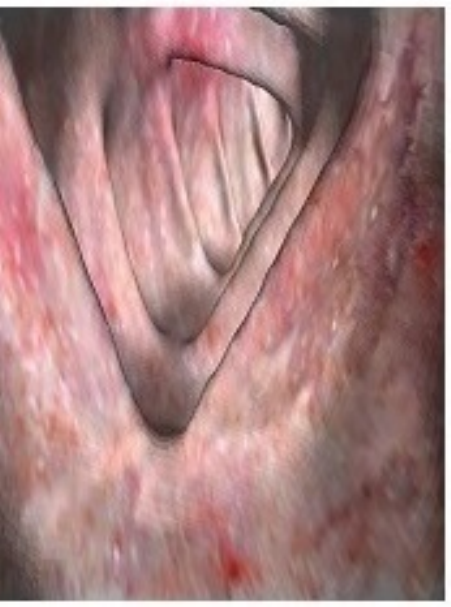}&
\includegraphics[width=0.23\textwidth, height=0.23\textwidth]{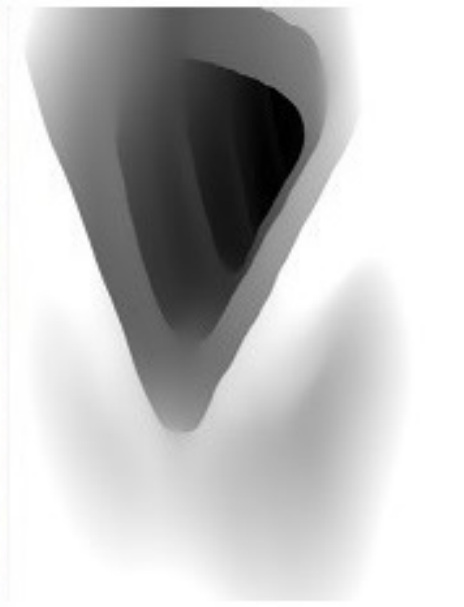}&
\includegraphics[width=0.23\textwidth, height=0.23\textwidth]{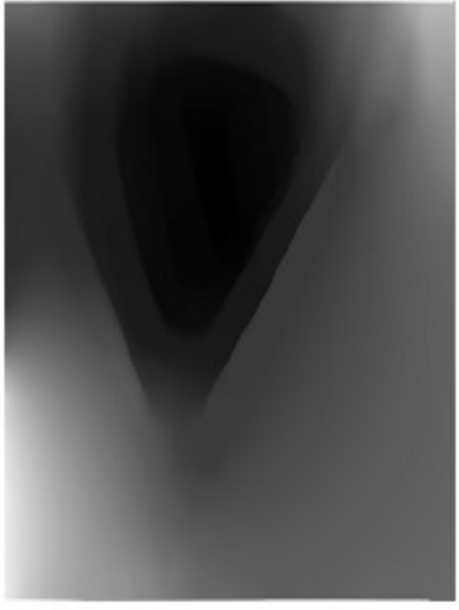}\\
(d) & (e) & (f)\\
\includegraphics[width=0.23\textwidth, height=0.23\textwidth]{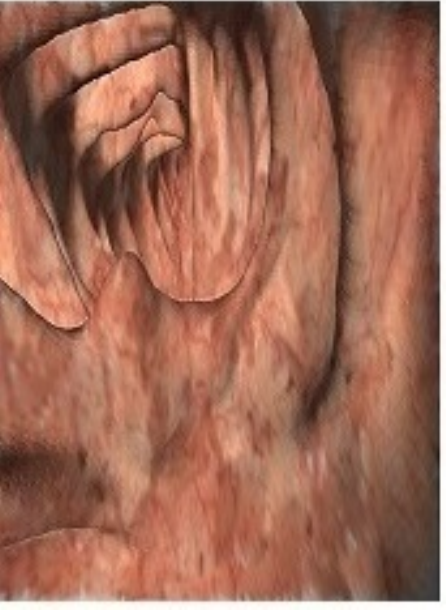}&
\includegraphics[width=0.23\textwidth, height=0.23\textwidth]{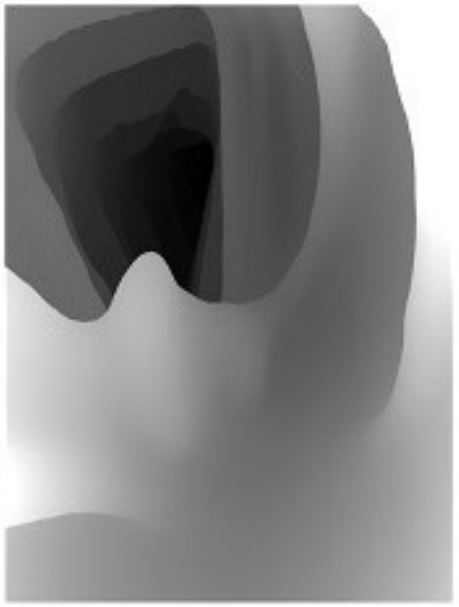}&
\includegraphics[width=0.23\textwidth, height=0.23\textwidth]{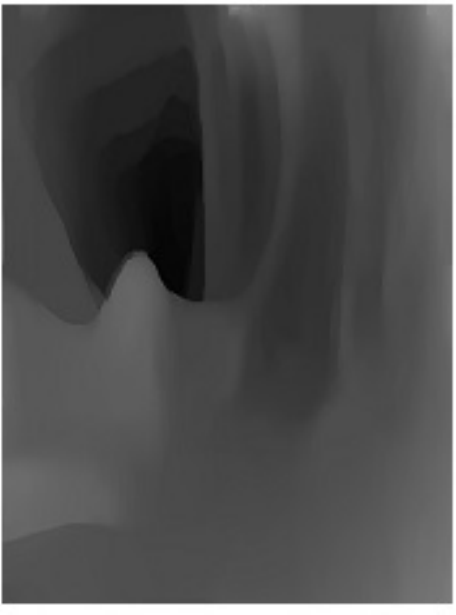}\\
(g) & (h) & (i)
\end{tabular}
\end{center}
\caption{The input image, true depth and the computed inferred depth with different texture mappings, (a) No Texture, (d) Texture 1, and (g) Texture 2.}
\label{fig:dicteval}
\end{figure}

We performed a 20-fold cross validation on our dictionary and show the error metric results in Table~\ref{tab:errordict}, with the ``No Texture'' label. We also collect 200 more images and their corresponding depth maps with two different texture mappings for testing purposes. The texture mappings, ``Texture 1'' and ``Texture 2'' are shown in Figure \ref{fig:dicteval}(d) and Figure \ref{fig:dicteval}(g). We train the dictionary and test it on these 100 ``Texture 1'' and 100 ``Texture 2'' images. The results for these texture mappings are shown in Table~\ref{tab:errordict}.

Denoting $x$ as estimated depth and $y$ as ground truth depth, we compute:

\begin{enumerate}
    \item Normalized root mean squared error (NRMSE):\\ $NRMSE = \frac{\sqrt{\frac{\sum_{i=1}^{n} (x_i-y_i)^2 }{n}}}{x_max - x_min}$. The closer to 0 this error, the more similar the two images.
    \item Hausdorff distance (HD):\\
        $H(x,y) = \max(h(x,y),h(y,x))$ with the directed Hausdorff distance defined as: $\vec{h}(x,y) = \underset{a\in x}{\max}\underset{b\in y}{\min}\|a-b\|$. The smaller the distance, the more similar the two images.
    \item Structural Similarity (SSIM) index: The SSIM metric is calculated on various windows of an image. The measure between two windows $x$ and $y$ of common size $N\times N$ is:\\
        $SSIM(x,y)=\frac{(2\mu_x \mu_y + c_1)(2\sigma_{xy}+c_2)}{(\mu_{x}^2 + \mu_{y}^2 + c_1)(\sigma_{x}^2 + \sigma_{y}^2 + c_2)}$\\
        where $\mu_x$ and $\mu_y$ are (respectively) the local sample means of $x$ and $y$, $\sigma_x$ and $\sigma_y$ are (respectively) the local sample standard deviations of $x$ and $y$, and $\sigma_{xy}$ is the sample cross correlation of $x$ and $y$ after removing their means. The items $c_1=(k_1 L)^2$ and $c_2=(k_2 L)^2$ are two variables to stabilize the division with weak denominator; $L$ is the dynamic range of the pixel values ($=2^{\#bits/pixel} - 1$) and $k_1=0.01$ and $k_2=0.03$ by default. The resultant SSIM index is a decimal value between -1 and 1, and value 1 indicates identical images. It is calculated on window sizes of $8\times8$. The window is displaced pixel-by-pixel on the image.
\end{enumerate}

\subsection{Distortion Agnostic}
\begin{figure}[h!]
\begin{center}
\begin{tabular}{cc}
\includegraphics[width=0.23\textwidth, height=0.23\textwidth]{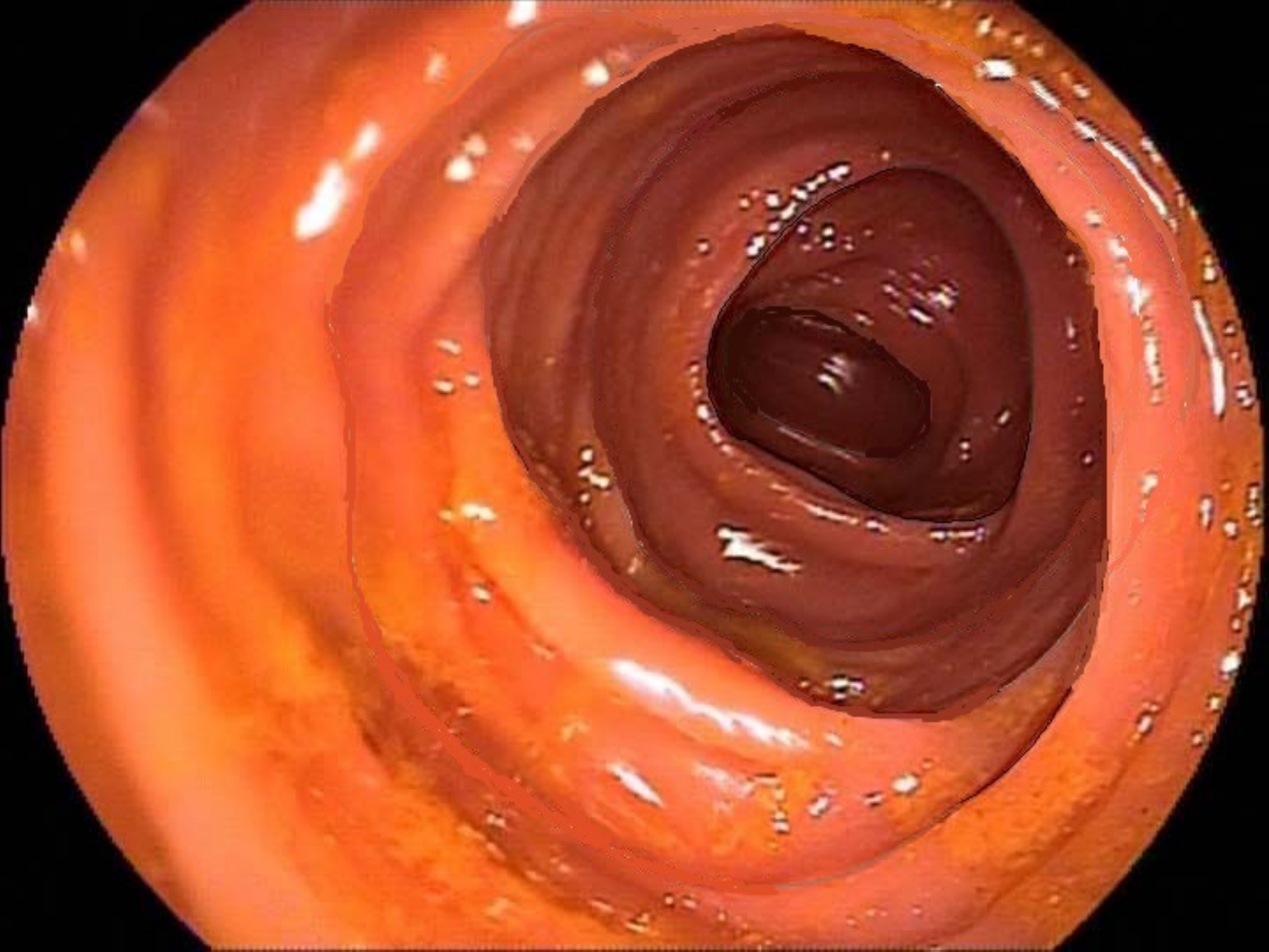}&
\includegraphics[width=0.23\textwidth, height=0.23\textwidth]{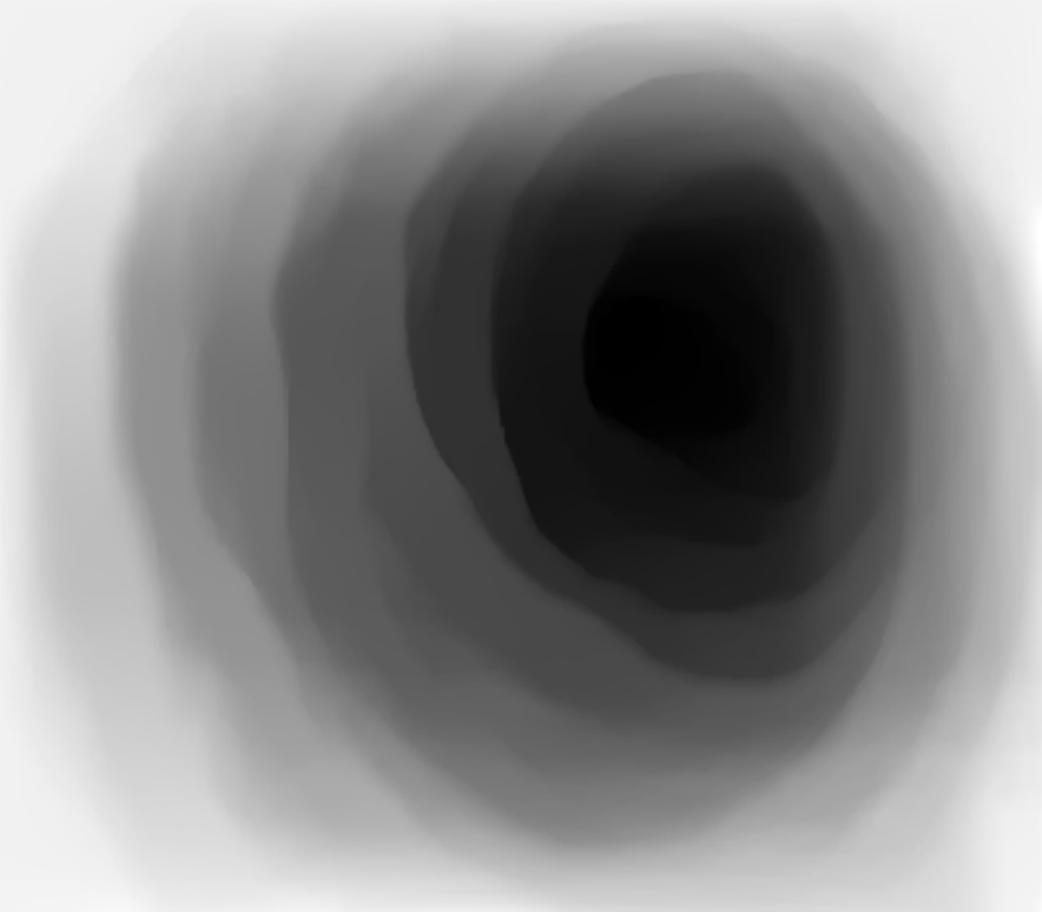}\\
(a) & (b)\\
\includegraphics[width=0.23\textwidth, height=0.23\textwidth]{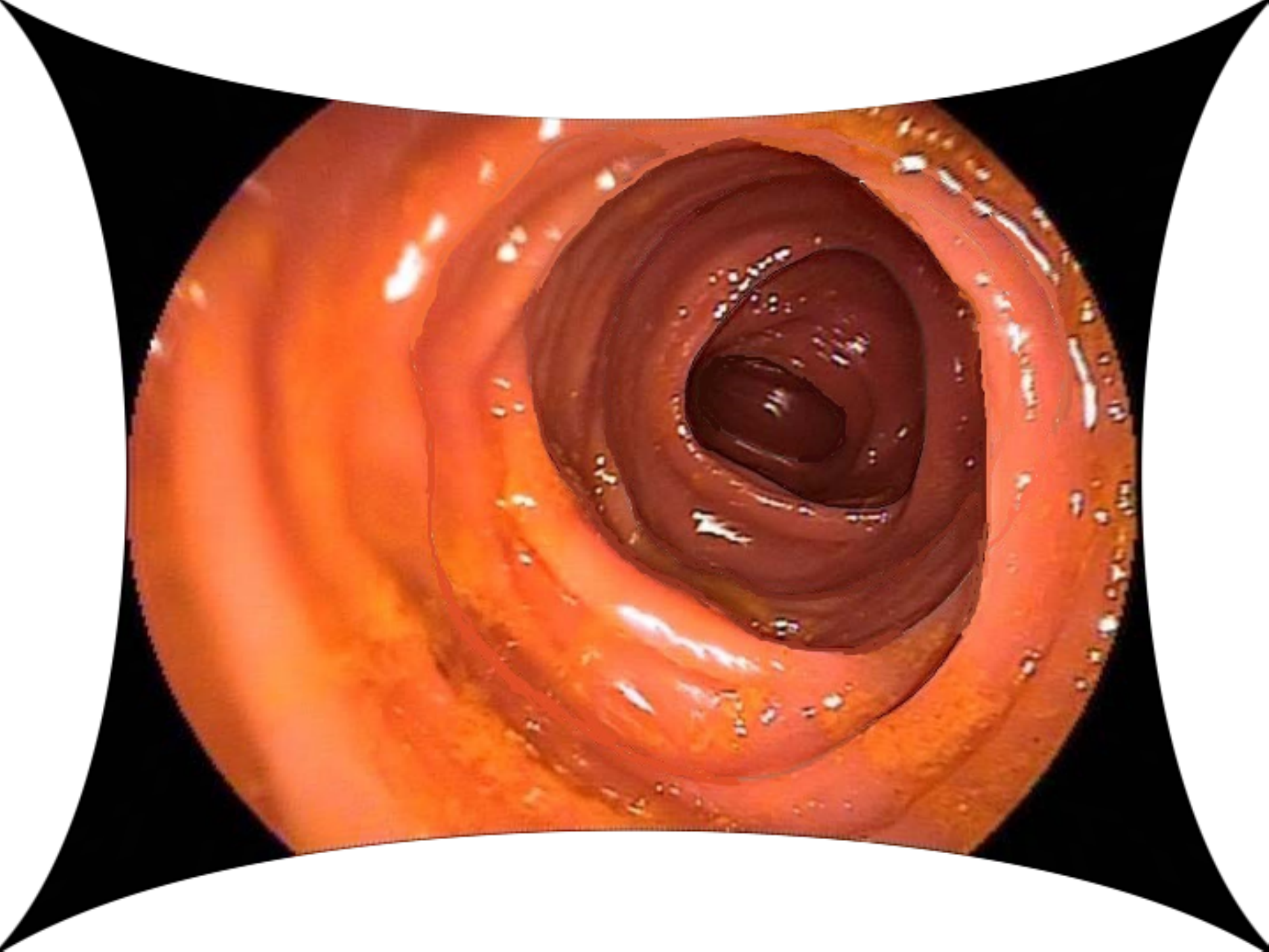}&
\includegraphics[width=0.23\textwidth, height=0.23\textwidth]{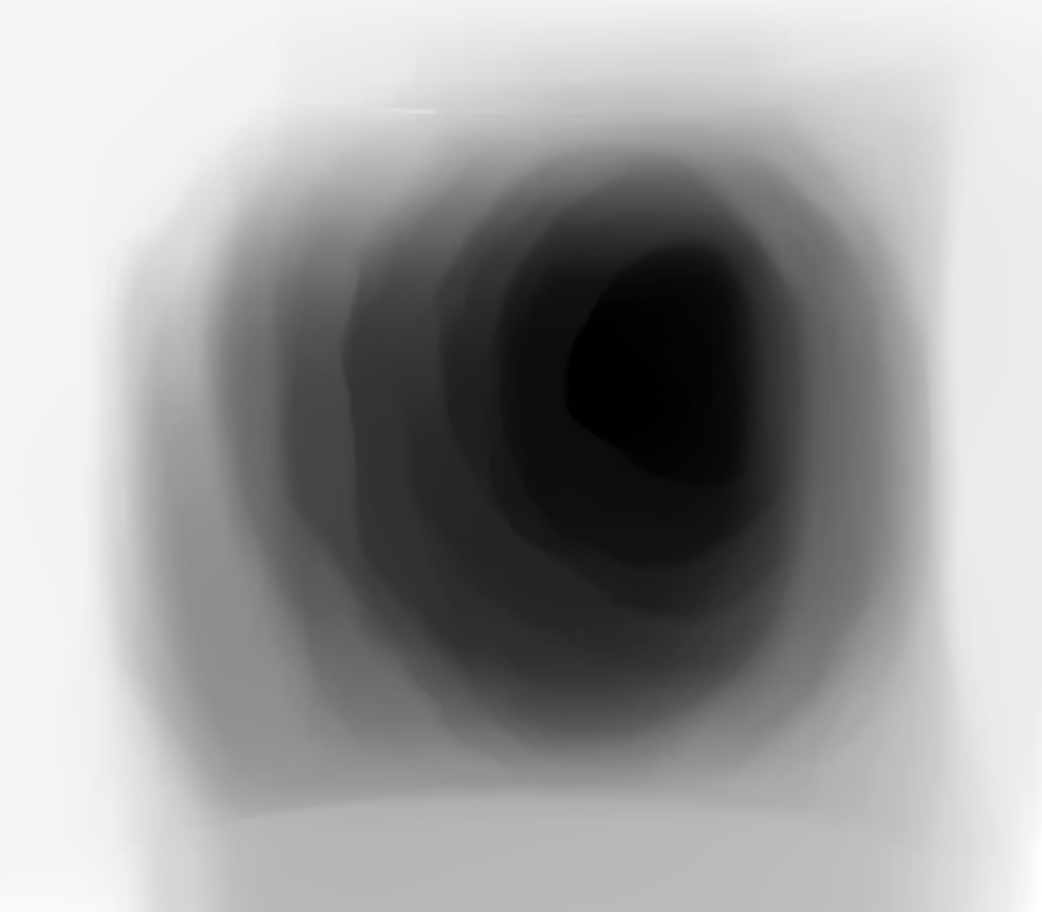}\\
(c) & (d)
\end{tabular}
\end{center}
\caption{(a) The fisheye distorted input image and (b) the corresponding depth output. (c) The undistorted input and (d) the corresponding depth output.}
\label{fig:undist}
\end{figure}

We also show that this approach is agnostic to lens distortion (such as fisheye distortion prevalent in the current endoscopes). The results for fisheye distortion and without fisheye distortion are shown in Figure~\ref{fig:undist}. In order to defish (remove fisheye distortion, Figure \ref{fig:undist}c), the given fisheye distorted input (Figure~\ref{fig:undist}a), we use the approach introduced by Marino et al. \cite{marino:2008a}.

\setlength{\tabcolsep}{4pt}

\subsection{Comparative Evaluation}
\begin{figure}[h!]
\begin{center}
\begin{tabular}{ccc}
Input frame & Hong et al. \cite{Hong:2014} & Our Result\\
\includegraphics[width=0.23\textwidth, height=0.23\textwidth]{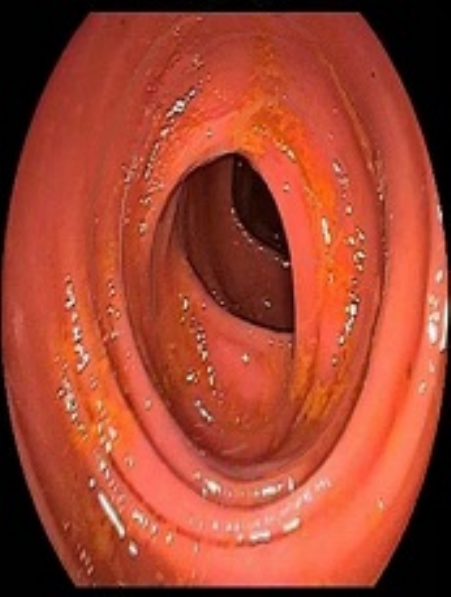}&
\includegraphics[width=0.23\textwidth, height=0.23\textwidth]{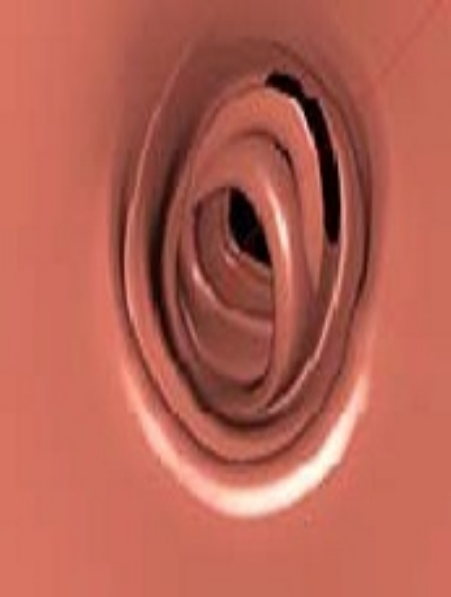}&
\includegraphics[width=0.23\textwidth, height=0.23\textwidth]{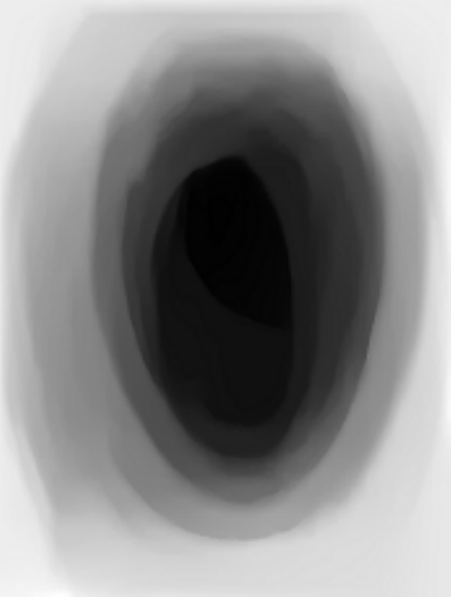}\\
(a) & (b) & (c)
\end{tabular}
\end{center}
\caption{(a) Input image, (b) the corresponding reconstruction from Hong et al. \cite{Hong:2014} and (c) our depth result.}
\label{fig:comparison}
\end{figure}

Hong et al. \cite{Hong:2014} extrapolate information which can give incorrect results as shown in Figure \ref{fig:comparison}. Moreover, their surface generation results are significantly distorted from the original image which makes it very hard to quantitatively evaluate their results. They assume a lot of things due to which it is hard to justify any metrics they estimate. Their algorithm discards the images which have a surface view of the colon.

\setlength{\tabcolsep}{4pt}

\subsection{Computer-Aided Detection Evaluation}
We tested our approach on 300 endoscopy video frames from patients with colon cancer, with polyp annotation or not, marked independently by two medical experts. 75 are polyp frames and 225 are normal frames. Some of the results are shown in Figure \ref{fig:polyps}. The confusion matrix in Table \ref{tab:conf} outlines the number of detections of polyp and normal frames. We obtained the best recall ($=TP/(TP+FN)$) of 84.0\% and the best specificity ($=TN/(FP+TN)$) value of 83.4\%.

\setlength{\tabcolsep}{4pt}
\begin{table}[ht!]
\caption{Error metrics for dictionary evaluation with Normalized Root Mean Squared Error (NRMSE), Hausdorff Distance (HD), and Structural Similarity Index (SSIM)}
\label{tab:errordict}
\begin{center}
\begin{tabular}{c||c|c|c}
\hline
Texture Mapping & NRMSE & HD & SSIM\\
\hline
No Texture & 0.57 & 0.56 & 0.35\\
Texture 1 & 0.49 & 0.44 & 0.31\\
Texture 2 & 0.43 & 0.43 & 0.30\\
\hline
\end{tabular}
\end{center}
\end{table}



\setlength{\tabcolsep}{2.2pt}
\begin{table}[ht!]
\caption{Confusion matrix of polyp frame detection using our computer-aided detection algorithm, where TP is True Positive, FP is False Positive, TN is True Negative, and FN is False Negative.}
\label{tab:conf}
\begin{center}
\begin{tabular}{c|c|c}
\hline
Actual/Predicted & Polyp & Normal\\
\hline
Polyp & TP (63) & FN (12)\\
Normal & FP (37) & TN (188)\\
\hline
\end{tabular}
\end{center}
\end{table}

\setlength{\tabcolsep}{4pt}
\begin{figure}[h]
\begin{center}
\begin{tabular}{cc}
Input Frame & Depth View\\
\includegraphics[width=0.23\textwidth, height=0.23\textwidth]{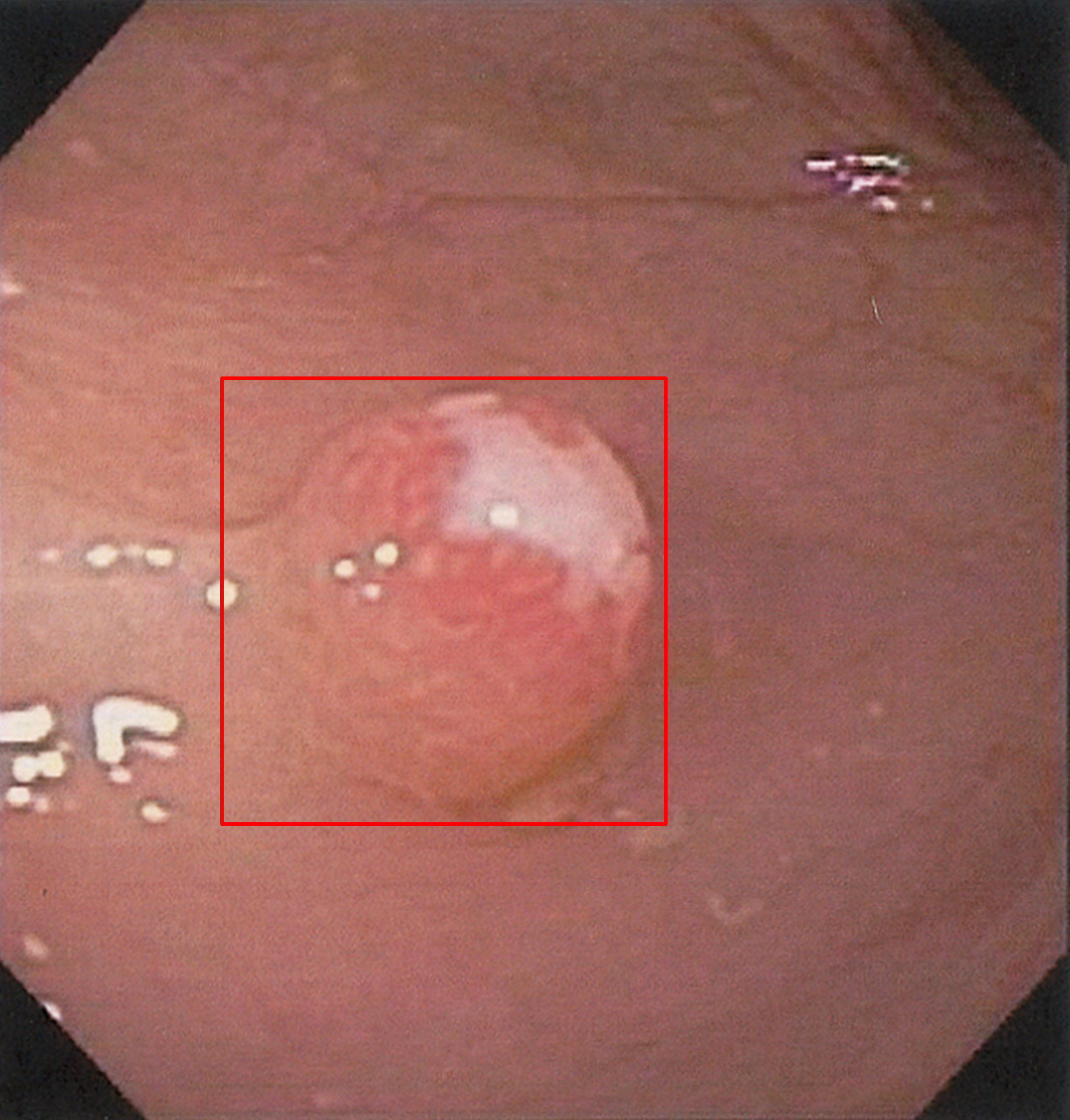}&
\includegraphics[width=0.23\textwidth, height=0.23\textwidth]{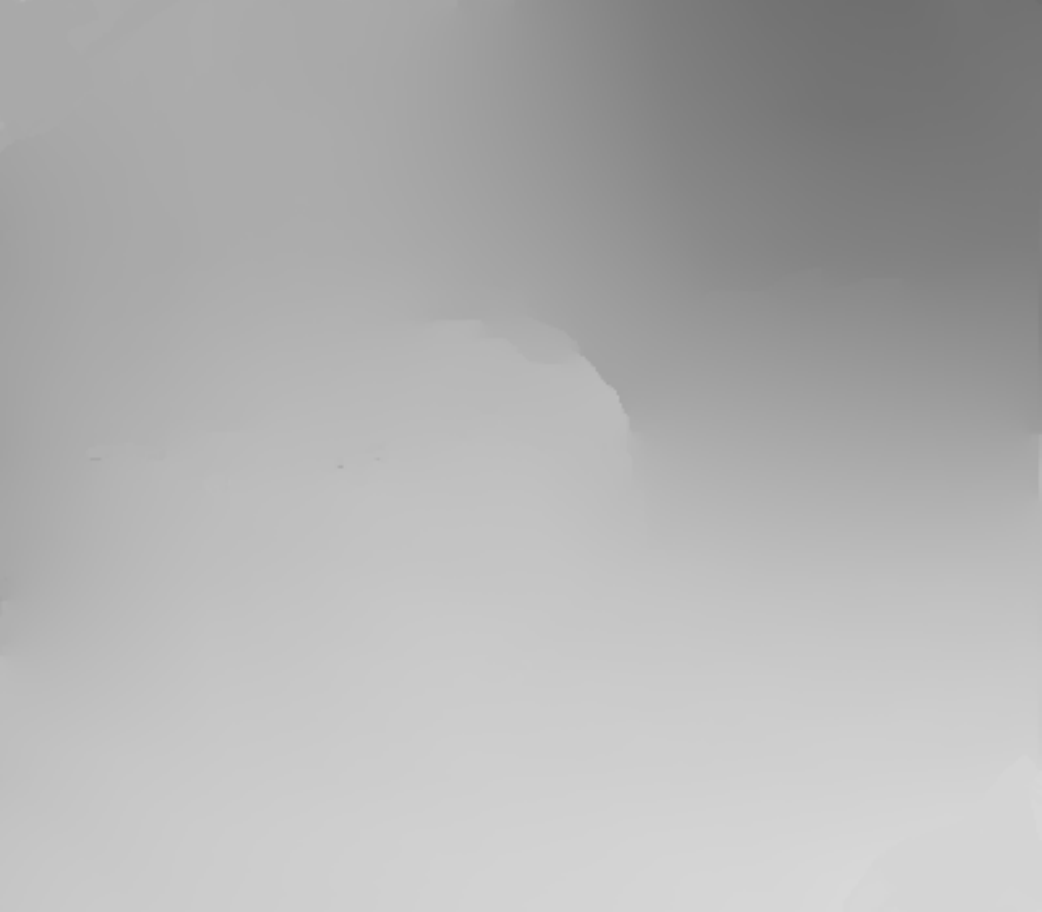}\\
\includegraphics[width=0.23\textwidth, height=0.23\textwidth]{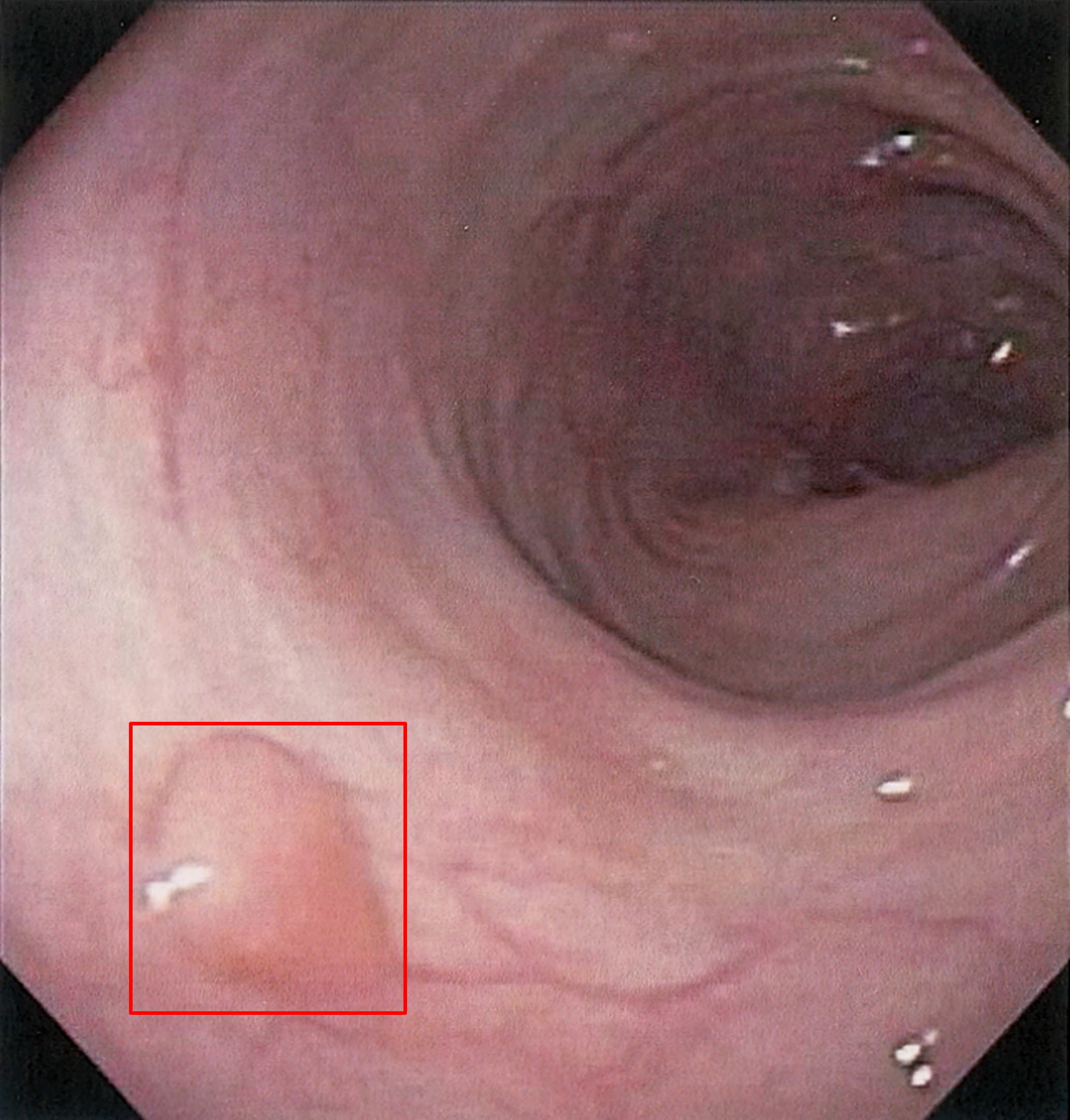}&
\includegraphics[width=0.23\textwidth, height=0.23\textwidth]{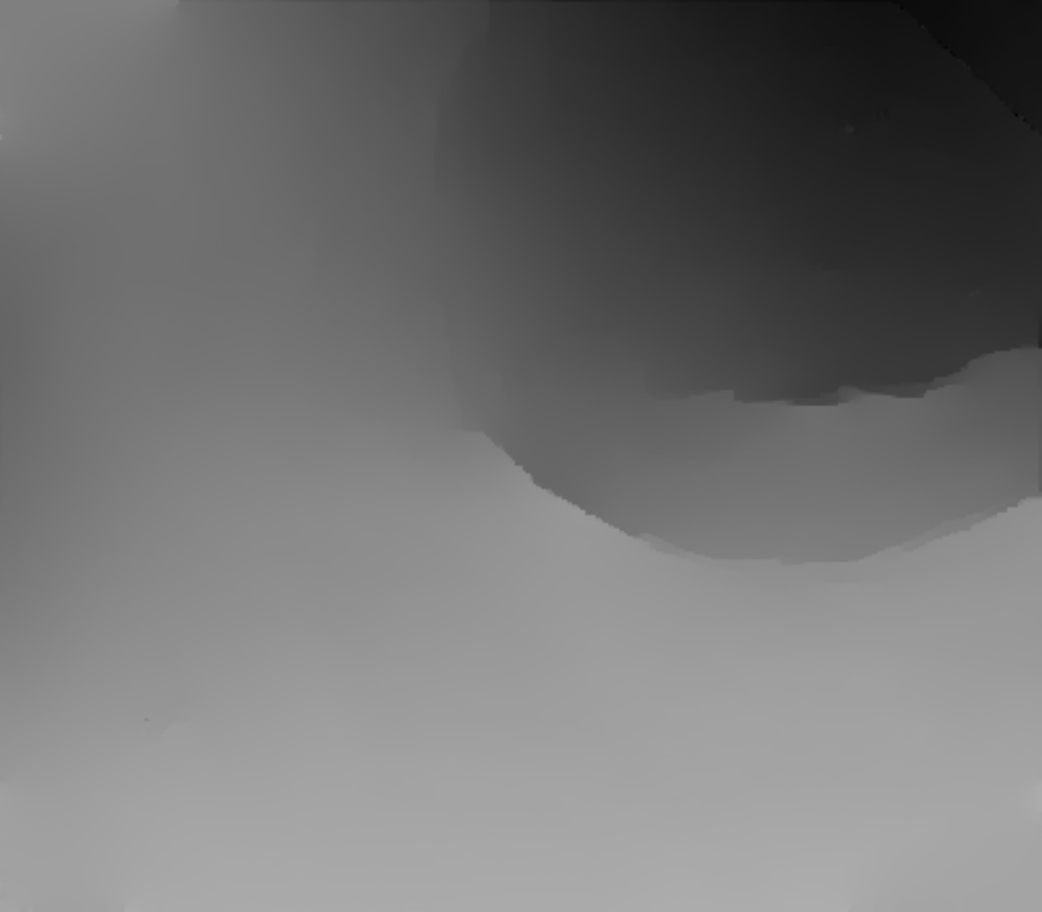}
\end{tabular}
\end{center}
\caption{Failure case. The polyp is not detected because of the similarity between the polyp texture and the background. The polyp is identified by a red bounding box.}
\label{fig:fail}
\end{figure}

\setlength{\tabcolsep}{4pt}
\begin{figure}[h]
\begin{center}
\begin{tabular}{cc}
\includegraphics[width=0.23\textwidth, height=0.23\textwidth]{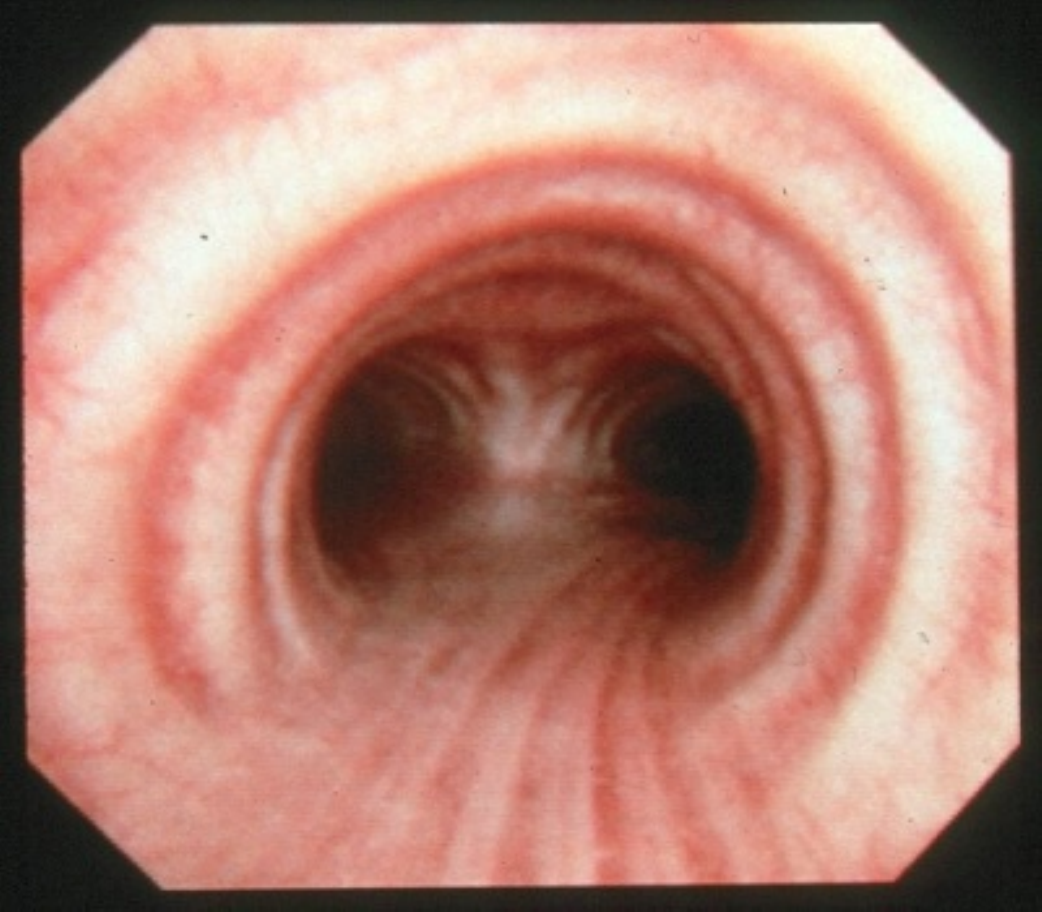}&
\includegraphics[width=0.23\textwidth, height=0.23\textwidth]{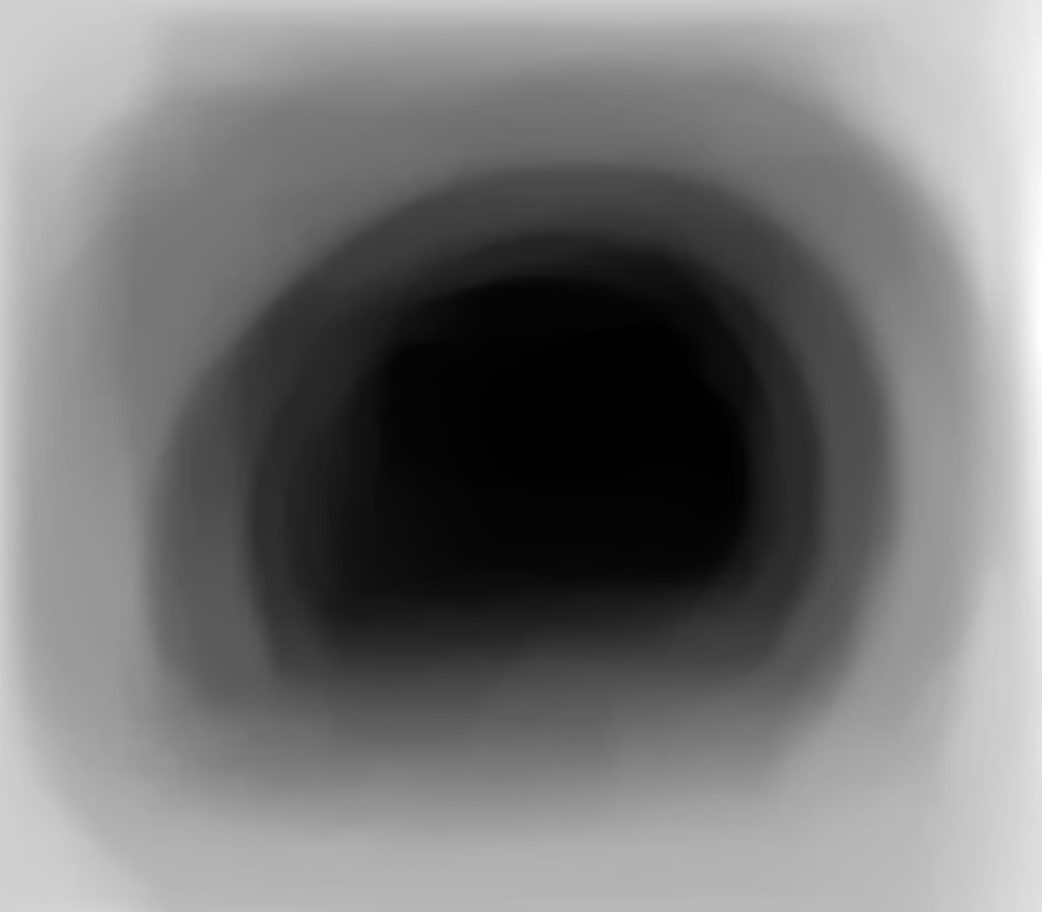}
\end{tabular}
\end{center}
\caption{Broncho Trachea. Input frame and the depth view, using the same dictionary created for the colonoscopy video frames.}
\label{fig:broncho}
\end{figure}

\section{DISCUSSION}
\label{sec:discussion}
To give an objective view of our methodology, we present results on some input OC images, used in Hong et al. \cite{Hong:2009}. Apart from these images, the other colonoscopy images used in this work are taken from the publicly available NIBIB Image and Clinical Data Repository provided by the NIH.

Some of the computed results for polyps, mucosa and lumen views dense depth map representations are presented in Figure \ref{fig:results1}. Without taking any prior assumptions on the intrinsic parameters of the camera, we are able to recover dense representation for a given video frame, which makes our approach more robust.

Our method still has a number of limitations. First, our framework focuses on the relative depth for our CAD algorithm; the absolute depth is not critical for the current work. Second, the performance is not a concern at the moment, but in the future we will improve upon this aspect. Moreover, we are still in the process of improving our estimation algorithm in terms of accuracy. Some of the failure cases are shown in Figure~\ref{fig:fail}. We will explore the use of approximated nearest neighbor search, file indexing, and deep learning approaches with augmented training datasets to improve our overall accuracy and performance.

Since our dictionary captures dense representation of the structures seen by the endoscope, we can extend our approach to other endoscopy procedures, such as bronchoscopy, cystoscopy, etc. A result for a bronchoscopy video frame is shown in Figure~\ref{fig:broncho}.

\section{CONCLUSION}
We presented an algorithm to automatically detect polyps in a colonoscopy image. We used a machine learning algorithm based on virtual colonoscopy training datasets to first infer depth maps for optical colonoscopy images and then use different layers of depth to compute and delineate the polyp boundaries in an image. We were able to achieve best recall of 84.0\% and the specificity value of 83.4\%. In the future, we will extend our method to other types of endoscopies such as cystoscopy, bronchoscopy, nasopharyngoscopy, etc. We also plan to jointly recover depth and normal information for a given video frame and propagate this information to the successive frames for more accurate reconstruction of a given set of successive video frames.

\section{ACKNOWLEDGEMENTS}
We would like to thank Dr. James Brief and Dr. Jeffrey Morganstern at the Department of Gastroenterology, Stony Brook University Hospital, for their help and guidance with this project.


\end{document}